\documentclass{article}

\usepackage[main, final]{neurips_2026}

\usepackage[utf8]{inputenc} % allow utf-8 input
\usepackage[T1]{fontenc}    % use 8-bit T1 fonts
\usepackage{hyperref}       % hyperlinks
\usepackage{url}            % simple URL typesetting
\usepackage{amsfonts}       % blackboard math symbols
\usepackage{nicefrac}       % compact symbols for 1/2, etc.
\usepackage{microtype}      % microtypography

\usepackage{amsmath}
\usepackage{amssymb}
\usepackage{mathtools}
\usepackage{amsthm}

\usepackage[table]{xcolor}
\usepackage{multirow}
\usepackage{makecell}
\usepackage{pifont}       
\usepackage{bbding}      
\usepackage{fontawesome}
\usepackage{subcaption}
\usepackage{arydshln}
\usepackage{tabularx}
\usepackage{mdframed}
\usepackage[most]{tcolorbox}
\usepackage{arydshln}
\usepackage{adjustbox}
\usepackage{wrapfig}
\usepackage{titletoc}

\usepackage{booktabs,array,colortbl,graphicx}

\definecolor{table-blue}{RGB}{173, 216, 230}
\definecolor{row-highlight}{RGB}{220, 230, 242}
\definecolor{qwen-header}{RGB}{235, 240, 248}
\definecolor{llama-header}{RGB}{248, 244, 235}
\definecolor{section-gray}{RGB}{245, 245, 245}
\definecolor{group-header}{RGB}{230, 235, 245}

\definecolor{zhz_gray}{rgb}{0.8,0.8,0.8}
\definecolor{darkgreen}{rgb}{0.0, 0.5, 0.0} % Custom dark green color
\definecolor{darkred}{rgb}{0.5, 0.0, 0.0}   % Custom dark red color

\definecolor{row-alt}{RGB}{248, 250, 253}      % very light cool tint
\definecolor{hl-llama}{RGB}{234, 242, 255}     % soft blue
\definecolor{hl-qwen}{RGB}{236, 250, 240}      % soft green

\title{MemSkill: Learning and Evolving Memory Skills for Self-Evolving Agents}

\author{%
\textbf{Haozhen Zhang}$^{1}$ \quad
\textbf{Quanyu Long}$^{1}$ \quad
\textbf{Jianzhu Bao}$^{1}$ \quad
\textbf{Tao Feng}$^{2}$ \\
\textbf{Weizhi Zhang}$^{3}$ \quad
\textbf{Haodong Yue}$^{4}$ \quad
\textbf{Wenya Wang}$^{1}$ \\
$^{1}$Nanyang Technological University \quad
$^{2}$University of Illinois Urbana-Champaign \\
$^{3}$University of Illinois Chicago \quad
$^{4}$Tsinghua University \\
\texttt{haozhen001@e.ntu.edu.sg} \quad
\texttt{wangwy@ntu.edu.sg}
}

\begin{document}

\maketitle

\begin{abstract}

Most Large Language Model (LLM) agent memory systems rely on a small set of static, hand-designed operations for extracting memory. These fixed procedures hard-code human priors about what to store and how to revise memory, making them rigid under diverse interaction patterns and inefficient on long histories.
To this end, we present \textbf{MemSkill}, which reframes these operations as learnable and evolvable memory skills, structured and reusable routines for extracting, consolidating, and pruning information from interaction traces.
Inspired by the design philosophy of agent skills, MemSkill employs a \emph{controller} that learns to select a small set of relevant skills, paired with an LLM-based \emph{executor} that produces skill-guided memories.
Beyond learning skill selection, MemSkill introduces a \emph{designer} that periodically reviews hard cases where selected skills yield incorrect or incomplete memories, and evolves the skill set by proposing refinements and new skills.
Together, MemSkill forms a closed-loop procedure that improves both the skill-selection policy and the skill set itself.
Experiments on LoCoMo, LongMemEval, HotpotQA, and ALFWorld demonstrate that MemSkill improves task performance over strong baselines and generalizes well across settings.
Further analyses shed light on how skills evolve, offering insights toward more adaptive, self-evolving memory management for LLM agents.
Code is available at \href{https://github.com/ViktorAxelsen/MemSkill}{https://github.com/ViktorAxelsen/MemSkill}

\end{abstract}

\section{Introduction}

As Large Language Model (LLM) agents engage in longer, open-ended interactions, they must handle growing histories that are essential yet challenging to leverage, motivating memory for retaining experience and maintaining coherence~\citep{hu2025memory}.
This need has driven rapid progress in agent memory, including approaches that summarize and retrieve past interactions or manage external memory stores~\citep{kang2025memory, chhikara2025mem0, packer2023memgpt, xu2025mem}. 
However, most methods still rely on static, hand-designed memory mechanisms, including fixed operation primitives (e.g., add/update/delete/skip)~\citep{wang2025mem, yan2025memory} and heuristic modules that govern what to store, how to revise it~\citep{kang2025memory, fang2025lightmem}, and when to prune it. Such designs bake in strong human assumptions and often suffer under diverse interaction patterns, scaling poorly as histories grow.

We argue that this formulation fundamentally limits the adaptability of agent memory.
Rather than treating memory as the output of fixed operations or hand-designed modules, we propose to elevate memory extraction itself into a \emph{learnable abstraction}.
Concretely, we view memory construction as the outcome of applying a small set of generic, reusable \emph{memory skills}: structured behaviors that specify when and how interaction traces should be transformed into memory and revised over time.
This perspective reveals a key bottleneck of prior pipelines: they hard-code memory behaviors into fixed procedural workflows that interleave heuristics with LLM-mediated extraction and revision, making them brittle under distribution shift~\citep{fang2025lightmem}.

Under this view, an ideal agent memory system should satisfy three properties.
(i) \textit{Minimal reliance on human priors.}
Instead of manually encoding what is worth remembering for a domain~\citep{zhong2024memorybank}, memory behaviors should be shaped by interaction data and updated as task demands evolve.
(ii) \textit{Support for larger extraction granularity.}
Many approaches are tuned to a fixed unit, such as per-turn processing~\citep{fang2025lightmem}, and can weaken when applied to longer spans.
A practical system should be able to operate at larger extraction granularity when needed.
(iii) \textit{Skill-conditioned, compositional memory construction.}
Existing systems often decompose memory construction into specialized modules~\citep{kang2025memory}.
In contrast, we prefer to \emph{select and compose} a small set of relevant skills for the current context and apply them in one generation step, enabling flexible reuse and evolution of memory behaviors.

\begin{figure*}[t]
	\centering
	\includegraphics[width=1.0\linewidth]{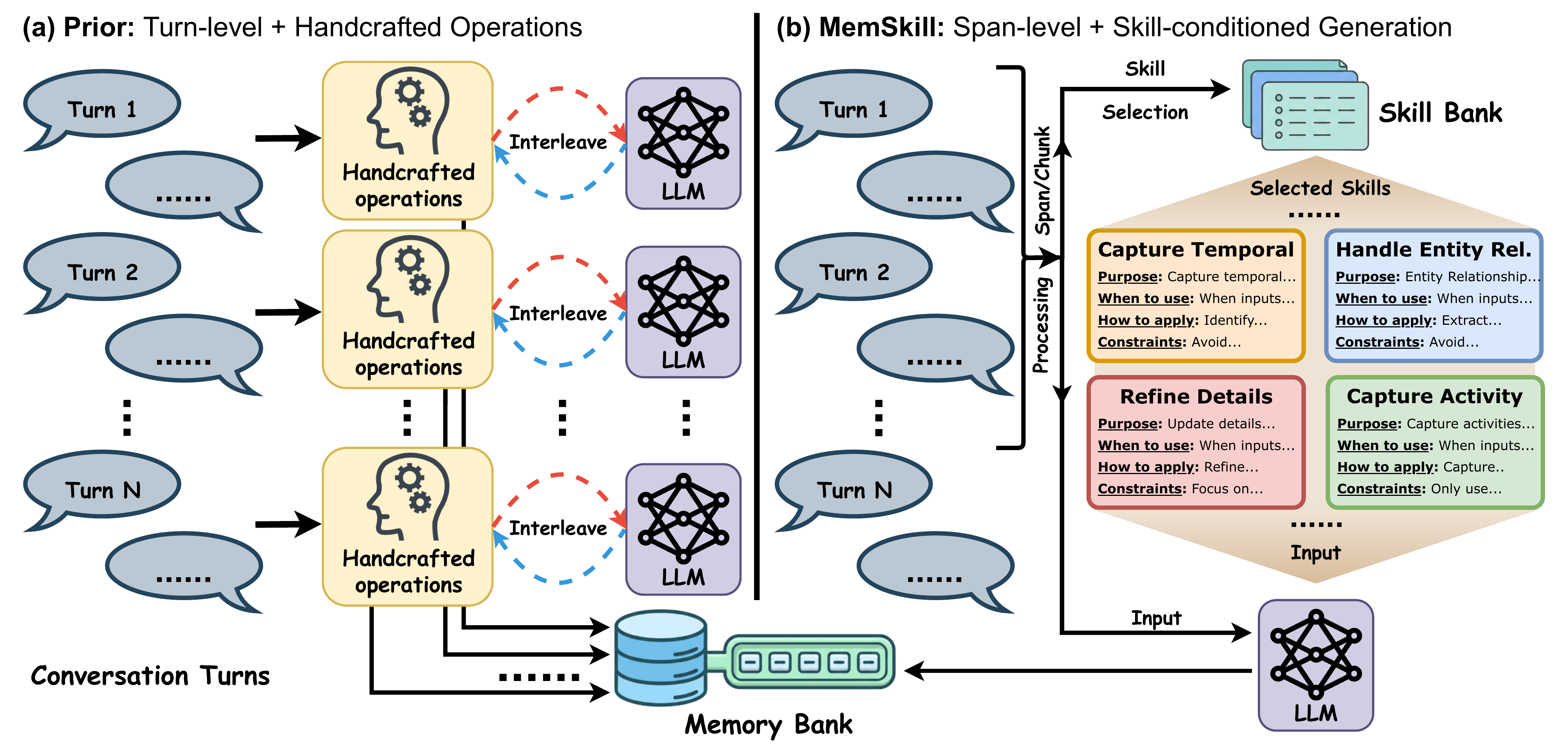}
        \caption{\textbf{Comparison between (a) prior turn-level, handcrafted operations and (b) MemSkill’s span-level, skill-conditioned generation.} Prior methods interleave handcrafted operations with LLM calls to incrementally extract and revise memory turn by turn, while MemSkill selects a small set of skills from a shared skill bank and applies them in one pass to produce skill-guided memories.}
	\label{intro_figure}
\end{figure*}

Based on the above observations, we introduce \textbf{MemSkill}, which reframes memory operations as a learnable and evolvable set of memory skills. 
MemSkill maintains a shared \emph{skill bank}, where each skill captures a reusable way to extract, consolidate, or revise memories from interaction text (Figure~\ref{intro_figure} shows the template of a memory skill). 
Given the current context, a \emph{controller} learns to select a small set of relevant skills, and an LLM-based \emph{executor} conditions on these skills to generate skill-guided memories in one pass.
This skill-conditioned formulation is not tied to a fixed extraction unit and can be applied to different span lengths when processing long interaction histories.

Crucially, MemSkill goes beyond learning how to use a fixed set of skills. We introduce a closed-loop evolution process that alternates between learning to use the current skill bank and evolving the skill bank itself. 
Specifically, we train the \emph{controller} with reinforcement learning (RL) using downstream task signals as feedback for skill selection. Periodically, a \emph{designer} aggregates the hardest cases produced during training, selects representative failures, and uses an LLM to refine existing skills and propose new ones. 
After each evolution step, the controller continues training on the evolved skill bank, with additional exploration to facilitate adopting newly introduced skills.
Overall, this process gradually strengthens both the skill selection policy and the evolving skill bank, moving toward a more adaptive memory management system driven by interaction data.

Experiments on LoCoMo, LongMemEval, HotpotQA, and ALFWorld show that MemSkill consistently improves task performance and generalizes well. Further analyses validate key components and showcase representative evolved skills, offering insights toward more adaptive, self-evolving memory management for LLM agents.
\textbf{Our contributions can be summarized as follows.}
\begin{itemize}
\item We propose \textbf{MemSkill}, an agent memory method that represents memory operations as an evolving skill bank, where each skill provides reusable guidance for selecting, extracting, and organizing useful memories. This turns memory construction from a fixed handcrafted pipeline into an adaptive skill-conditioned generation process.
\item We introduce a closed-loop optimization recipe that combines reinforcement learning for skill selection with LLM-guided skill evolution from hard cases, enabling continual refinement of the skill bank and taking a step toward self-evolving agent memory systems.
\item We evaluate MemSkill on LoCoMo, LongMemEval, HotpotQA, and ALFWorld, demonstrating consistent gains and strong transfer ability across conversational QA and embodied interaction settings, offering insights for self-evolving memory in LLM agents.
\end{itemize}

\section{Related Work}

\subsection{LLM Agent Memory Systems}

Prior work on agent memory focuses on constructing external memories from interaction histories and leveraging them to support downstream reasoning and decision making. 
Typical pipelines periodically extract salient information into a memory store, retrieve relevant entries for a new query, and update the store via consolidation or pruning~\citep{kang2025memory, zhong2024memorybank, xu2025mem, packer2023memgpt, chhikara2025mem0, fang2025lightmem}. More recently, learning-based approaches including Memory-R1~\citep{yan2025memory} and Mem-$\alpha$~\cite{wang2025mem} optimize memory management with reinforcement learning (RL) using downstream task signals. 
Similarly, AgeMem~\citep{yu2026agentic} employs RL to manage short- and long-term memory through predefined operations.
Despite this progress, memory management is still largely governed by static, hand-crafted routines for extraction, consolidation, and pruning.

Several concurrent works also explore self-evolving memory in agent settings, but differ from our focus. Evo-Memory~\citep{wei2025evo} evaluates streaming memory evolution, MemEvolve~\citep{zhang2025memevolve} optimizes predefined memory architectures, MemGen~\citep{zhang2025memgen} targets latent memory for reasoning, and ReasoningBank~\citep{ouyang2025reasoningbank} distills reasoning strategies from experience. 
By contrast, we target the evolution of memory skills themselves, enabling the system to refine and grow reusable memory operations over time.

\subsection{Self-Evolving LLM Agents}

Recent work on self-evolving LLM agents studies how agents can improve from interaction experience with minimal manual supervision. 
ExpeL~\citep{zhao2024expel} distills trajectories into editable natural-language insights and retrieves relevant experiences to guide future decisions, while EvolveR~\citep{wu2025evolver} formalizes an experience lifecycle that consolidates interactions into reusable principles and closes the loop with reinforcement learning updates. 
A complementary line reduces reliance on curated data via self-play style curricula: Absolute Zero Reasoner~\citep{zhao2025absolute} trains a proposer and solver with verifiable rewards from a code executor, and Multi-Agent Evolve~\citep{chen2025multi} extends this to a proposer solver judge triad with LLM-based evaluation; R-Zero~\citep{huang2025r} follows a similar challenger solver co-evolution pattern. 
Beyond curricula, systems such as AgentEvolver~\citep{zhai2025agentevolver} and RAGEN~\citep{wang2025ragen} study efficient agent learning dynamics and stabilization in multi-turn RL settings, while ADAS~\citep{hu2024automated} and AlphaEvolve~\citep{novikov2025alphaevolve} explore automated discovery and evolutionary improvement of agent designs. Finally, SkillWeaver~\citep{zheng2025skillweaver} shows that agents can discover and refine reusable skills for web interaction. In contrast, our focus is on self-evolving \emph{memory skills} that govern how agents construct and revise memories over time.

\section{Method}\label{sec:method}

In this section, we first overview MemSkill (Section~\ref{sec:method_overview}), followed by the \emph{skill bank} (Section~\ref{sec:skill_bank}) and three core components: \emph{controller} (Section~\ref{sec:controller}), \emph{executor} (Section~\ref{sec:executor}), and \emph{designer} (Section~\ref{sec:designer}). We then describe the closed-loop optimization of skill selection and evolution (Section~\ref{sec:closed_loop}).

\begin{figure*}[t]
	\centering
	\includegraphics[width=1.0\linewidth]{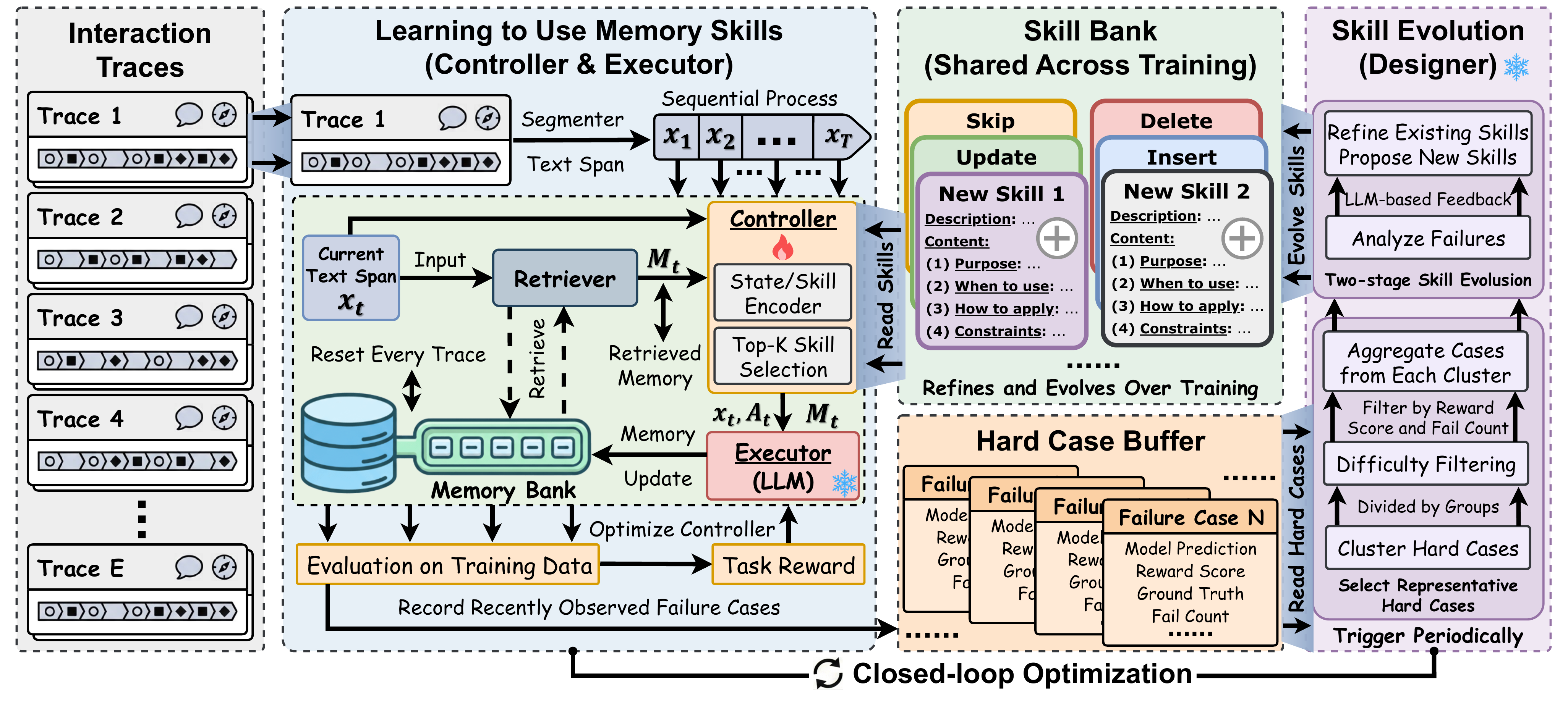}
        \caption{\textbf{MemSkill architecture overview.} Given an interaction trace, MemSkill processes it span by span: the controller selects a Top-$K$ subset of skills from a shared \emph{skill bank} conditioned on the current text span and retrieved memories, and an LLM executor applies the selected skills in one pass to update the trace-specific \emph{memory bank}. The constructed memory is then evaluated on memory-dependent training queries to provide task reward for optimizing the controller, while query-centric failures are logged into a sliding hard-case buffer. Periodically, the designer mines representative hard cases to refine existing skills and propose new ones, yielding alternating phases of skill usage and skill evolution. More skill case study can be found in Section~\ref{sec:case_study} and Appendix~\ref{appendix:case_study}.}
	\label{model_figure}
\end{figure*}

\subsection{Overview}\label{sec:method_overview}

As shown in Figure~\ref{model_figure}, we propose \textbf{MemSkill}, which optimizes agent memory through two intertwined processes. First, it \textbf{learns to use a given skill bank}: a controller selects context-relevant skills, and an executor applies them to produce memory updates. Second, it \textbf{improves the skill bank itself}: a designer periodically revises existing skills and adds new ones based on challenging training cases.

To disentangle trace-specific memories from reusable memory management knowledge, MemSkill maintains two stores. 
The \emph{memory bank} is trace-specific and stores memories for each training trace (e.g., a long dialogue). 
In contrast, the \emph{skill bank} is shared across traces and contains reusable memory skills.
During training, the controller and executor build each trace's memory bank, while the designer updates the shared skill bank between phases. This alternating procedure gradually improves both the skill selection policy and the skill bank for memory construction.

\subsection{Skill Bank}\label{sec:skill_bank}

As shown in Figure~\ref{model_figure}, a \emph{memory skill} specifies a reusable memory operation as structured guidance, including when it is applicable and how it should be applied to the current context. Concretely, each skill $s \in \mathcal{S}$ contains (i) a short \emph{description} for skill representation and selection, and (ii) a detailed \emph{content} specification that instructs the executor on memory extraction or revision.

We start from a minimal set of general-purpose primitives to ensure a stable and functional initialization. Specifically, we initialize the skill bank with four basic skills corresponding to canonical memory operations: \textsc{Insert}, \textsc{Update}, \textsc{Delete}, and \textsc{Skip}. Starting from this minimal set, the designer progressively refines existing skills and expands the bank by proposing new skills that address uncovered failure modes. (Appendix~\ref{appendix:case_study} details skill description)

\subsection{Learning to Use Memory Skills}\label{sec:learn_to_use_skills}

In this part, we describe how MemSkill learns to use memory skills, covering (i) the skill-selection policy and (ii) skill-conditioned memory construction.

\subsubsection{Controller: Skill Selection Policy}\label{sec:controller}

To enable effective skill selection as the \emph{skill bank} evolves, we introduce a controller that selects a small set of relevant memory skills for the current context. 
At each memory construction step, \textbf{we update memory at the span level}: we split each interaction trace (e.g., a dialogue) into fixed-length contiguous spans by token count and process them sequentially. For each span, the controller conditions its selection on (i) the current text span and (ii) the retrieved existing memories from the current trace's memory bank (empty for the initial span), rather than operating turn by turn.

To remain compatible with a variable-size skill bank as it continuously evolves, the controller scores each skill by measuring the semantic distance between the current state and skill representations, naturally supporting a changing skill set while staying sensitive to what is already stored in memory.

\textbf{State and skill representations.}
Formally, let $x_t$ denote the current text span at step $t$, and let $M_t=\{m_{t,1},\dots,m_{t,R}\}$ be the retrieved memories from the current trace's memory bank. We first encode $x_t$ and $M_t$ with a fixed embedding model $e(\cdot)$, aggregate the retrieved memory embeddings by element-wise averaging, and concatenate it with the span embedding:
\begin{equation}
h_t = f_{\theta}^{\mathrm{ctx}}\!\left([e(x_t);\bar{m}_t]\right), \qquad
u_i = f_{\theta}^{\mathrm{skill}}\!\left(e(\mathrm{desc}(s_i))\right),
\end{equation}
where $\bar{m}_t=\frac{1}{R}\sum_{r=1}^{R}e(m_{t,r})$ denotes the aggregated memory embedding, and is omitted when $M_t$ is empty. Here, $u_i$ is the representation of skill $s_i \in \mathcal{S}_t$, computed from its description as a compact and stable semantic signal rather than the full skill content. The embedding model $e(\cdot)$ is shared and fixed, while $f_{\theta}^{\mathrm{ctx}}$ and $f_{\theta}^{\mathrm{skill}}$ are trainable neural networks in the controller for learnable skill selection.

\textbf{Compatibility with an evolving skill bank.}
Instead of producing a fixed-dimensional action head tied to a fixed number of skills, the controller concatenates the state representation with each candidate skill representation and applies a shared scorer to all such state-skill pairs in parallel:
\begin{equation}
z_{t,i} = f_{\theta}^{\mathrm{score}}\!\left([h_t;u_i]\right), \qquad
p_\theta(i \mid h_t) = \mathrm{softmax}(z_t)_i,
\end{equation}
where $f_{\theta}^{\mathrm{score}}$ is a trainable neural network, and $z_t \in \mathbb{R}^{|\mathcal{S}_t|}$ adapts as the skill bank evolves.

\textbf{Top-$K$ skill selection.}
Given the categorical distribution $p_\theta(i\mid h_t)$ over the current skill bank $\mathcal{S}_t$, the controller selects an ordered Top-$K$ set of skills
$A_t=(a_{t,1},\dots,a_{t,K})$ (e.g., via Gumbel-Top-$K$~\citep{kool2019stochastic}), and only passes the selected skills to the executor, keeping the skill context concise and relevant.

\subsubsection{Executor: Skill-Conditioned Memory Extraction}\label{sec:executor}

Given the selected skills $A_t$, the fixed executor constructs memory updates by conditioning an LLM on (i) the current text span $x_t$, (ii) the retrieved memory items $M_t$, and (iii) the selected skills $A_t$. This mirrors skill-conditioned inference in agent systems, where a small set of relevant skills is provided to guide behavior for the current context.
The executor produces structured memory updates, which are parsed and applied to the trace's memory bank. 
By composing several skills for the same text span and extracting memory in one LLM call, MemSkill reduces repeated per-turn processing and scales better to long interaction histories.
Appendix~\ref{appendix:prompts} details the executor prompt.

\subsubsection{Controller Optimization}\label{sec:controller_optimization}

We train the controller with reinforcement learning, using downstream task performance as feedback for its skill selections. For each training trace, the controller makes a sequence of Top-$K$ selections while the executor incrementally builds the trace-specific memory bank. After construction, we evaluate the resulting memory bank on the trace's memory-dependent training queries and use the resulting task performance as the reward (e.g., F1 or success rate).

A key technical detail is that the controller's action is an ordered Top-$K$ \emph{list} selected without replacement, rather than a single discrete action. We therefore compute the joint log-probability $\log \pi_\theta(A_t\mid h_t)$ under this selection process and use it in standard policy-gradient objectives~\citep{schulman2017proximal} via importance weighting and clipping. Concretely, for $A_t=(a_{t,1},\dots,a_{t,K})$, the joint probability is
\begin{equation}
\pi_\theta(A_t \mid h_t)
= \prod_{j=1}^{K}
\frac{p_\theta(a_{t,j}\mid h_t)}
{1-\sum_{\ell<j} p_\theta(a_{t,\ell}\mid h_t)} ,
\label{eq:topk_joint_prob_main}
\end{equation}
which reduces to the single-action case when $K=1$. Appendix~\ref{appendix:rl_objective} gives implementation details.

\subsection{Skill Evolution through Designer Feedback}\label{sec:designer}

Beyond learning to select from a fixed set of skills, MemSkill evolves the skill bank using an LLM-based designer (fixed) that operates periodically during training.

\textbf{Hard-case buffer.} During controller training, we maintain a sliding-window buffer of challenging cases observed recently. Each case is query-centric, recording the query along with its ground-truth and metadata (e.g., retrieved memories and model prediction), as well as summary statistics such as task performance and the number of failures observed so far.
The buffer uses two expiration rules: cases are removed if they become too old (exceeding a maximum training step gap) or if the buffer reaches its capacity limit, which tracks recent failure patterns without growing unbounded.

\textbf{Selecting representative hard cases.} To focus designer updates on impactful failures, we cluster cases (e.g., KMeans) into groups that naturally reflect different query or error types. Within each cluster, we prioritize representative cases using a difficulty score that increases when task performance is low and when the same case fails repeatedly. This produces a compact set of high-value cases for skill evolution while preserving diversity across error types.

\textbf{Two-stage skill evolution.} The designer updates the skill bank in two stages. First, it employs an LLM to analyze the selected hard cases and identify what memory behaviors are missing or mis-specified. Second, it uses the resulting analysis to propose concrete edits to existing skills and to introduce new skills. We keep the designer description concise and provide prompts in Appendix~\ref{appendix:prompts}.

Notably, we maintain snapshots of the best-performing skill bank and roll back if an update degrades performance, with early stopping when repeated designer updates fail to improve the training signal. After each evolution step, we also briefly increase exploration by biasing selection toward newly introduced skills, encouraging the controller to try them and facilitating efficient learning of their utility.
Due to page limit, more details about the designer can be found in Appendix~\ref{appendix:detail_component}.

\subsection{Closed-Loop Optimization}\label{sec:closed_loop}

MemSkill alternates between (i) learning to select and apply skills to build memory banks and (ii) evolving the skill bank from hard cases mined during training. Each cycle begins with controller training on the current skill bank, where the executor constructs memories and accumulates challenging cases. The designer then updates the skill bank using representative hard cases, optionally rolling back to a prior snapshot if performance regresses. The next cycle resumes controller training on the updated skill bank, with additional exploration to encourage early use of new skills. Over cycles, this closed loop gradually improves how skills are selected, applied, and refined for memory construction.

\section{Experiments}
\label{sec:exp}

\subsection{Experiment Setup}
\label{sec:exp_setup}

\textbf{Datasets and Baselines.}
We evaluate MemSkill on four benchmarks: LoCoMo~\citep{maharana2024evaluating-locomo}, LongMemEval~\citep{wu2024longmemeval}, HotpotQA~\cite{yang2018hotpotqa}, and ALFWorld~\cite{shridhar2020alfworld}, where HotpotQA is used in Section~\ref{sec:exp_shift} to study skill transfer under distribution shift. The remaining three benchmarks cover two representative settings. 
(i) \textit{Conversational Benchmarks} include LoCoMo and LongMemEval, which evaluate memory construction from long, dialogue-style interaction histories. 
For these datasets, we report F1-score (F1) and an LLM-based judge score (L-J).
(ii) \textit{Embodied Interactive Tasks} are evaluated on ALFWorld with two standard subsets, ALF-Seen and ALF-Unseen, and we report success rate (SR) and the number of environment interaction steps (\#Stps).
Appendix~\ref{appendix:exp_eval} provides dataset splits.

We compare MemSkill against several strong baselines: (1) \textbf{No-Memory}, which answers directly without an external memory; (2) \textbf{CoN}~\citep{yu2024chain}; (3) \textbf{ReadAgent}~\citep{lee2024human}; (4) \textbf{MemoryBank}~\citep{zhong2024memorybank}; (5) \textbf{A-MEM}~\citep{xu2025mem}; (6) \textbf{Mem0}~\citep{chhikara2025mem0}; (7) \textbf{LangMem}~\citep{LangMem}; and (8) \textbf{MemoryOS}~\citep{kang2025memory}.
Overall, this setup spans diverse benchmarks and baselines, enabling a broad and consistent comparison across diverse settings.

\textbf{Implementation Details.}
We instantiate $f_{\theta}^{\mathrm{ctx}}$, $f_{\theta}^{\mathrm{skill}}$, and $f_{\theta}^{\mathrm{score}}$ as independent lightweight multilayer perceptrons (MLPs), and use LLaMA-3.3-70B-Instruct~\citep{grattafiori2024llama} and Qwen3-Next-80B-A3B-Instruct~\citep{yang2025qwen3} as the base LLMs, accessed through an API service. Unless otherwise specified, we train MemSkill on LLaMA and use Qwen only for transfer experiments. LongMemEval is also evaluated in a transfer setting, where we directly apply the skills learned on LoCoMo without further training.

During training, we optimize the controller with PPO~\cite{schulman2017proximal}. MemSkill constructs memory at the span level; on conversational benchmarks, each dialogue session is a processing unit, and the controller selects $K{=}3$ skills per unit. We instantiate $e(\cdot)$ with Qwen3-Embedding-0.6B~\citep{yang2025qwen3}, also used as the memory retriever, and retrieve up to 20 memory items for MemSkill and all baselines for consistency. For the designer, skill evolution is triggered every 100 training steps, with at most 3 skill edits per evolution round. For ALFWorld, we cap environment interactions at 50 steps.

At evaluation time, we keep the same span-level formulation and set the span/chunk size to 512 by default, while keeping the overall procedure unchanged. Unless otherwise specified, we use $K{=}7$ skills for LoCoMo and LongMemEval at evaluation time, and $K{=}5$ for ALFWorld. Additional implementation details and prompt templates are provided in Appendix~\ref{appendix:imple_detail} and Appendix~\ref{appendix:prompts}.

\begin{table*}[t]
    \footnotesize
    \centering
    \setlength{\tabcolsep}{5.5pt}
    \caption{\textbf{Main comparison results on LoCoMo, LongMemEval, and ALFWorld.}}
    \label{tab:exp_main}

    \begin{tabular}{>{\centering\arraybackslash}m{0.5cm} l cc cc c | cc cc c}
        \toprule
        \multirow{3}{*}{\textbf{Model}} &
        \multirow{3}{*}{\textbf{Methods}} &
        \multicolumn{5}{c}{\textbf{Conversational Benchmarks}} &
        \multicolumn{5}{c}{\textbf{Embodied Interactive Tasks}} \\
        \cmidrule(lr){3-7} \cmidrule(lr){8-12}
        & &
        \multicolumn{2}{c}{\textbf{LoCoMo}} &
        \multicolumn{2}{c}{\textbf{$^\blacktriangle$LongMemEval}} &
        \multicolumn{1}{c}{\textbf{Avg.}} &
        \multicolumn{2}{c}{\textbf{ALF-Seen$^\dagger$}} &
        \multicolumn{2}{c}{\textbf{ALF-Unseen$^\dagger$}} &
        \multicolumn{1}{c}{\textbf{Avg.}} \\
        \cmidrule(lr){3-4} \cmidrule(lr){5-6} \cmidrule(lr){7-7}
        \cmidrule(lr){8-9} \cmidrule(lr){10-11} \cmidrule(lr){12-12}
        & &
        \textbf{F1} & \textbf{L-J} &
        \textbf{F1} & \textbf{L-J} &
        \textbf{L-J} &
        \textbf{SR} & \textbf{\#Stps$\downarrow$} &
        \textbf{SR} & \textbf{\#Stps$\downarrow$} &
        \textbf{SR} \\
        \midrule

        % ===================== Base model block 1 =====================
        \multirow{9}{*}{\rotatebox[origin=c]{90}{\shortstack{\textbf{LLaMA3.3}\\\textbf{70B-Instruct}}}} &
        No-Memory         & - & - & - & - & - & 62.14 & 26.10 & 73.88 & 21.36 & 68.01 \\
        & CoN       & 30.86 & 41.72 & 30.78 & 56.44 & 49.08 & 75.00 & 19.15 & 80.60 & 17.38 & 77.80 \\
        & ReadAgent  & 28.63 & 38.25 & 24.48 & 42.62 & 40.44 & 62.86 & 26.14 & 71.64 & 22.88 & 67.25 \\
        & MemoryBank & 36.80 & 44.43 & 30.56 & 41.96 & 43.20 & 60.71 & 28.23 & 66.42 & 24.64 & 63.57 \\
        & A-MEM      & 39.39 & 49.71 & 25.83 & 38.04 & 43.88 & 62.86 & 27.53 & 70.15 & 23.79 & 66.51 \\
        & Mem0       & 25.48 & 34.58 & 30.25 & 46.81 & 40.70 & 74.29 & 19.77 & 81.34 & 17.15 & 77.82 \\
        & LangMem    & 30.91 & 35.82 & 18.36 & 24.35 & 30.09 & 72.86 & 21.25 & 79.85 & 18.30 & 76.36 \\
        & MemoryOS   & 41.39 & 48.64 & 17.59 & 39.83 & 44.24 & 57.86 & 27.94 & 65.67 & 24.46 & 61.77 \\
        & \cellcolor{hl-llama}\textbf{MemSkill} &
        \cellcolor{hl-llama}\textbf{44.21} & \cellcolor{hl-llama}\textbf{53.82} & \cellcolor{hl-llama}\textbf{31.12} & \cellcolor{hl-llama}\textbf{60.89} &
        \cellcolor{hl-llama}\textbf{57.36} &
        \cellcolor{hl-llama}\textbf{77.14} & \cellcolor{hl-llama}\textbf{18.91} &
        \cellcolor{hl-llama}\textbf{83.58} & \cellcolor{hl-llama}\textbf{16.63} &
        \cellcolor{hl-llama}\textbf{80.36} \\
        \midrule

        % ===================== Base model block 2 =====================
        \multirow{9}{*}{\rotatebox[origin=c]{90}{\shortstack{$^\blacktriangle$\textbf{Qwen3-Next}\\\textbf{80B-A3B-Instruct}}}} &
        No-Memory         & - & - & - & - & - & 63.57 & 24.61 & 60.45 & 26.57 & 62.01 \\
        & CoN       & 38.46 & 50.96 & \textbf{29.19} & 44.06 & 47.51 & 77.14 & 17.59 & 70.90 & 20.74 & 74.02 \\
        & ReadAgent  & 25.89 & 34.26 & 24.13 & 42.25 & 38.26 & 73.57 & 20.21 & 65.67 & 23.28 & 69.62 \\
        & MemoryBank & 29.56 & 44.15 & 8.45 & 26.37 & 35.26 & 63.57 & 23.68 & 52.24 & 29.01 & 57.91 \\
        & A-MEM      & 36.43 & 50.30 & 13.84 & 36.59 & 43.45 & 55.71 & 27.42 & 54.48 & 28.60 & 55.10 \\
        & Mem0       & 23.29 & 33.68 & 27.36 & 46.20 & 39.94 & 71.43 & 19.89 & 64.93 & 23.32 & 68.18 \\
        & LangMem    & 28.17 & 32.94 & 18.35 & 23.86 & 28.40 & 73.57 & 19.76 & 64.18 & 23.58 & 68.88 \\
        & MemoryOS   & 39.86 & 47.37 & 15.97 & 39.25 & 43.31 & 62.14 & 25.64 & 50.75 & 30.35 & 56.45 \\
        & \cellcolor{hl-qwen}\textbf{MemSkill} &
        \cellcolor{hl-qwen}\textbf{42.08} & \cellcolor{hl-qwen}\textbf{54.14} & \cellcolor{hl-qwen}25.29 & \cellcolor{hl-qwen}\textbf{60.40} &
        \cellcolor{hl-qwen}\textbf{57.27} &
        \cellcolor{hl-qwen}\textbf{85.71} & \cellcolor{hl-qwen}\textbf{13.84} &
        \cellcolor{hl-qwen}\textbf{76.87} & \cellcolor{hl-qwen}\textbf{18.16} &
        \cellcolor{hl-qwen}\textbf{81.29} \\
        \bottomrule

        \multicolumn{12}{l}{\small \textbf{Bold} indicates the best score within each base model block; $^\blacktriangle$ indicates transfer evaluation only.} \\
        \multicolumn{12}{l}{\small $^\dagger$ indicates evaluation with in-context demonstrations. Appendix~\ref{appendix:more_exp} reports more baselines and datasets.} \\
        \\
    \end{tabular}
    % \vspace{-4pt}
\end{table*}

\subsection{Comparison Experiments}
\label{sec:exp_main}

\textbf{Effectiveness across conversational and embodied settings.}
Table~\ref{tab:exp_main} summarizes the main comparison results on LoCoMo, LongMemEval, and ALFWorld. Across these datasets, MemSkill achieves the strongest overall performance among all compared methods. On conversational benchmarks, MemSkill attains the best LLM-judge scores on both LoCoMo and LongMemEval within each base-model block, indicating higher-quality constructed memories. 
In comparison, prior methods such as MemoryBank, A-MEM, and MemoryOS use fixed, manually specified memory procedures for extraction and revision, whereas MemSkill learns and evolves its skills from interaction, enabling better adaptation across contexts.
On ALFWorld, MemSkill achieves the highest success rates on both seen and unseen splits, indicating that skill-guided memory construction can benefit interactive decision making, whereas other baselines are less reliable at leveraging memory to support long-horizon action execution.
Overall, the results show that MemSkill is effective across diverse settings.

\textbf{Generalization across base models.}
A key advantage of MemSkill is strong generalization across base models. We train MemSkill only with LLaMA and directly transfer the learned skills to Qwen without retraining. Despite this strict transfer setting, MemSkill remains competitive and outperforms strong baselines on both conversational and embodied evaluations, showing that the evolved skills capture reusable memory behaviors that can be instantiated by different underlying LLMs.

\textbf{Cross-dataset transfer.}
MemSkill also generalizes across datasets within the same broad setting. In particular, LongMemEval is evaluated purely by transferring the skill bank learned on LoCoMo, yet MemSkill achieves the best results among all methods, suggesting that the learned skills are not overfit to a single benchmark. Section~\ref{sec:exp_shift} studies transfer under more pronounced distribution shifts.

\begin{figure*}[t]
	\centering
	\includegraphics[width=1.0\linewidth]{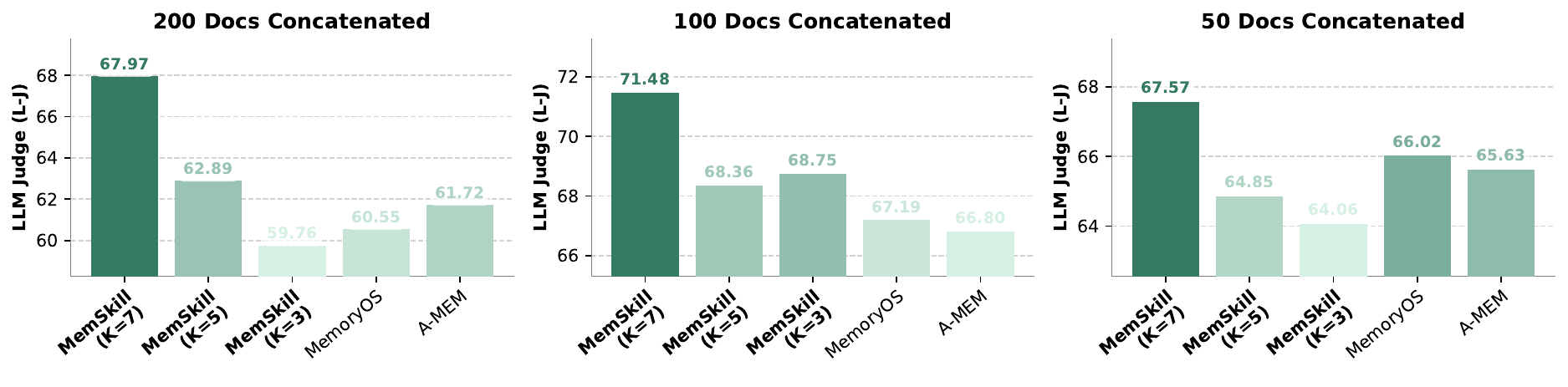}
        \caption{\textbf{Skill generalization under distribution shift on HotpotQA.} We transfer the LoCoMo-trained skill bank to HotpotQA and evaluate three context-length settings (50/100/200 concatenated documents) following~\citep{yu2025memagent}. Bars show LLM-judge (L-J) under LLaMA with different Top-$K$ skill counts, compared to MemoryOS and A-MEM.}
	\label{fig:hotpot_shift}
\end{figure*}

\subsection{Skill Generalization Under Distribution Shift}
\label{sec:exp_shift}

Beyond transfer within dialogue-style memory benchmarks, we evaluate whether learned skills generalize under a distribution shift in interaction format and evidence structure. 
Concretely, we directly apply the skill bank trained on LoCoMo to HotpotQA, where inputs are long-form, document-style narratives rather than multi-turn dialogues. 
Following the protocol in~\citep{yu2025memagent}, we test three context-length settings with increasing difficulty, corresponding to different numbers of concatenated documents (i.e., 50/100/200). 
All results in this section use LLaMA as the base model and report the LLM-judge score (L-J). 
For baselines, we include MemoryOS and A-MEM, the most competitive methods on conversational benchmarks in Table~\ref{tab:exp_main}, and omit weaker alternatives for clarity.

Figure~\ref{fig:hotpot_shift} shows that MemSkill transfers strongly to HotpotQA across all three context sizes. 
MemSkill consistently outperforms strong baselines such as MemoryOS and A-MEM, with gains becoming more pronounced in the challenging long-context setting.
These results suggest that the learned memory skills are not tied to dialogue-specific surface forms, but capture reusable extraction and revision behaviors that remain effective as input structure and retrieval demands change.

The same plots also reveal mild sensitivity to the number of selected skills $K$. Increasing $K$ generally improves performance, with $K{=}7$ achieving the best results across all three settings, while smaller $K$ can under-utilize the skill bank under longer contexts. Overall, the trend indicates that MemSkill benefits from composing multiple skills when the context becomes longer and noisier, while still maintaining strong transfer without any HotpotQA-specific training.

\subsection{Ablation Study}\label{sec:ablation}

We perform ablations to disentangle the contributions of (i) learning to select skills and (ii) evolving the skill bank. Table~\ref{tab:ablation_locomo} reports LLM Judge (L-J) results on LoCoMo under both base models (LLaMA and Qwen). 
As shown, \textbf{w/o controller (w/o Ctrl)} replaces the learned controller with random skill selection while keeping the rest of the pipeline unchanged. 
\textbf{w/o designer (w/o Des)} disables the designer and fixes the skill bank to the four initial primitives. 
\textbf{Refine-only (Ref.-only)} allows the designer to refine existing skills but prohibits adding new ones.

\begin{wraptable}{r}{0.35\columnwidth}
    \vspace{-1.0em}
    \footnotesize
    \centering
    \caption{\textbf{LoCoMo ablation} (L-J).}
    \label{tab:ablation_locomo}
    \setlength{\tabcolsep}{6pt}
    \small
    \begin{tabular}{lcc}
        \toprule
        \textbf{Variant} & \textbf{LLaMA} & \textbf{Qwen} \\
        \midrule
        \textbf{MemSkill} & \textbf{53.82} & \textbf{54.14} \\
        \quad w/o Ctrl & 48.43 & 42.84 \\
        \quad w/o Des & 46.50 & 36.15 \\
        \quad Ref.-only & 47.45 & 48.88 \\
        \bottomrule
    \end{tabular}
    \vspace{-0.7em}
\end{wraptable}

Across both base models, removing either component consistently degrades performance, confirming that MemSkill benefits from both targeted skill selection and skill evolution. 
In particular, random skill selection leads to a clear drop from the default setting, highlighting the importance of learning to choose relevant skills rather than providing arbitrary ones. 
Disabling the designer yields an even larger degradation, especially under Qwen, suggesting that evolving the skill bank is important for learning reusable memory behaviors that generalize beyond a fixed, manually specified operation set. 
Finally, refinement-only consistently outperforms static skills on both LLaMA and Qwen, with a particularly large gain under Qwen, yet remains below the default setting, indicating that introducing new skills yields additional benefits beyond refining the initial primitives.

\begin{figure*}[t]
	\centering
	\includegraphics[width=1.0\linewidth]{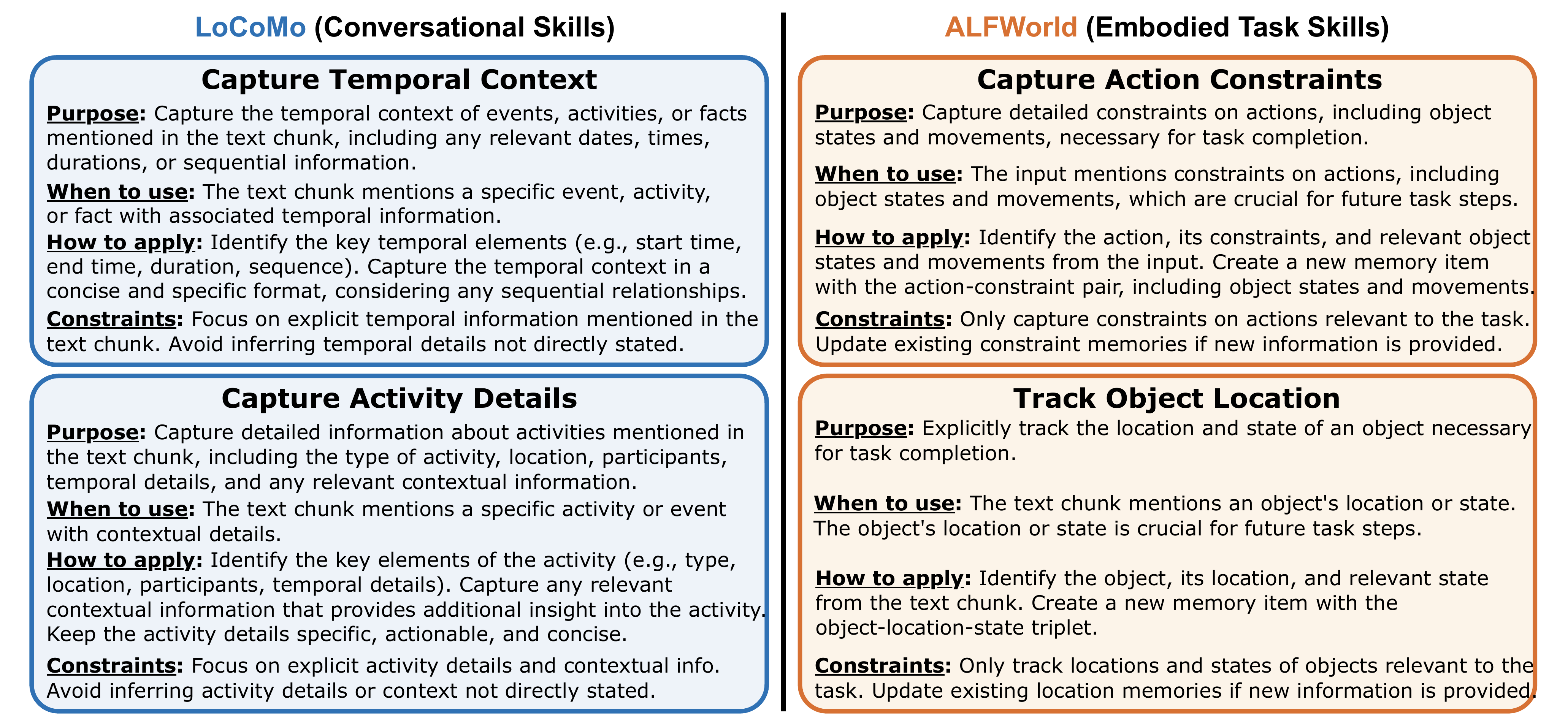}
        \caption{\textbf{Case study.} We show representative evolved skills learned on LoCoMo and ALFWorld.}
	\label{fig:case_study}
\end{figure*}

\subsection{Discussion}
\label{sec:case_study}

\textbf{Case study.}
To make MemSkill more interpretable, we inspect the final evolved skill bank and report representative skills from LoCoMo and ALFWorld. 
As shown in Figure~\ref{fig:case_study}, the learned skills show clear domain specialization. 
LoCoMo skills emphasize temporal context and activity details, suggesting that dialogue memory benefits from lightweight event structure. In contrast, ALFWorld skills focus on action constraints and object locations, showing that embodied tasks require actionable world-state memories for multi-step execution.
Overall, the evolved skill bank reflects recurring information needs from the data, rather than a fixed notion of what to remember.

Together, these skills show that MemSkill distills and refines reusable memory behaviors from interaction data, reducing reliance on hand-crafted memory designs. Appendix~\ref{appendix:case_study} gives more examples.

\textbf{Cost analysis.}
We conduct a runtime cost analysis on LoCoMo using LLaMA, additionally including \textbf{LightMem}~\citep{fang2025lightmem} as a baseline. We report L-J, input/output tokens, and LLM calls, accounting for all inference-time LLM calls from memory extraction and query answering, excluding LLM-judge calls used only for evaluation. Since MemSkill learns and evolves the skill bank before deployment, this preparation cost is amortized over repeated use, with further evolution triggered only occasionally. We therefore focus on runtime cost as the practical efficiency measure. At inference time, MemSkill constructs memory at the span level rather than turn by turn, substantially reducing LLM calls. To show this trade-off, we vary the span size (SS) and report the quality-cost frontier in Table~\ref{tab:cost_analysis_chunk}.

\begin{wraptable}{r}{0.51\columnwidth}
    \vspace{-1.0em}
    \footnotesize
    \centering
    \caption{\textbf{Cost analysis on LoCoMo.}}
    \label{tab:cost_analysis_chunk}
    \setlength{\tabcolsep}{6pt}
    \begin{tabular}{lcccc}
        \toprule
        \textbf{Setting} & \textbf{L-J} & \textbf{In (K)} & \textbf{Out (K)} & \textbf{Calls} \\
        \midrule
        MemoryOS  & 48.64 & 1013 & 165 & 1288 \\
        A-MEM     & 49.71 & 2850 & 362 & 1548 \\
        LightMem  & 51.95 & 789  & 209 & 685  \\
        \midrule
        \cellcolor{hl-llama}\textbf{Ours} (SS=128)  & \cellcolor{hl-llama}53.14 & \cellcolor{hl-llama}622 & \cellcolor{hl-llama}57 & \cellcolor{hl-llama}376 \\
        \cellcolor{hl-llama}\textbf{Ours} (SS=256)  & \cellcolor{hl-llama}51.61 & \cellcolor{hl-llama}390 & \cellcolor{hl-llama}19 & \cellcolor{hl-llama}270 \\
        \cellcolor{hl-llama}\textbf{Ours} (SS=512)  & \cellcolor{hl-llama}53.82 & \cellcolor{hl-llama}249 & \cellcolor{hl-llama}18 & \cellcolor{hl-llama}215 \\
        \cellcolor{hl-llama}\textbf{Ours} (SS=1024) & \cellcolor{hl-llama}48.11 & \cellcolor{hl-llama}188 & \cellcolor{hl-llama}11 & \cellcolor{hl-llama}187 \\
        \cellcolor{hl-llama}\textbf{Ours} (SS=2048) & \cellcolor{hl-llama}50.46 & \cellcolor{hl-llama}178 & \cellcolor{hl-llama}10 & \cellcolor{hl-llama}185 \\
        \bottomrule
    \end{tabular}
    \vspace{-0.7em}
\end{wraptable}

MemSkill achieves a stronger quality-cost trade-off than prior baselines. With moderate span sizes, it obtains higher L-J scores while using fewer input/output tokens and LLM calls. In particular, Span Size=512 offers the best overall balance, achieving the highest quality with much lower runtime cost than MemoryOS, A-MEM, and LightMem. This suggests span-level construction reduces redundant LLM calls without sacrificing memory quality. 
Non-monotonic performance reflects a granularity trade-off: smaller spans may fragment events, while larger spans may overlook details. Span size thus balances quality and efficiency.

\section{Conclusion}\label{sec:conclusion}

We present \textbf{MemSkill}, an agent memory method that reframes memory operations as an evolving skill bank. MemSkill learns to select relevant skills for each context span and conditions an LLM executor on them to construct skill-guided memories. Beyond learning to use a fixed skill set, MemSkill introduces a designer that improves the skill bank by refining existing skills and proposing new ones from challenging cases, forming a closed-loop training procedure. Experiments on LoCoMo, LongMemEval, HotpotQA, and ALFWorld show consistent improvements over strong baselines, while qualitative analyses illustrate how evolving skills enable more adaptive memory management. We hope MemSkill encourages future work on self-improving agent memory systems that learn not only to use memory, but also to continually improve how memory is constructed and maintained.

\begin{ack}

This research is supported by the National Research Foundation, Singapore under its National Cybersecurity R\&D Programme (CRPO-GC5-NTU-011). Any opinions, findings and conclusions or recommendations expressed in this material are those of the authors and do not reflect the views of National Research Foundation, Singapore.

\end{ack}

\bibliographystyle{plain}
\bibliography{ref}

%%%%%%%%%%%%%%%%%%%%%%%%%%%%%%%%%%%%%%%%%%%%%%%%%%%%%%%%%%%%
% \clearpage
\appendix

\startcontents[appendix]
\section*{Appendix Contents}
\printcontents[appendix]{}{1}{\setcounter{tocdepth}{2}}

% \clearpage

\section{More Experimental Results}
\label{appendix:more_exp}

\subsection{More Comparison Experiments}
\label{appendix:more_exp_comparison}

We further compare MemSkill with additional baselines on ALFWorld, including LightMem~\citep{fang2025lightmem}, AWM~\citep{wang2024agent}, and Expel~\citep{zhao2024expel}. As shown in Table~\ref{tab:more_exp_comparison}, MemSkill consistently achieves the best performance across both base models and evaluation splits. With LLaMA, MemSkill obtains an average success rate of 80.36, outperforming the strongest baseline LightMem by 5.53 points. The gain is especially clear on ALF-Unseen, where MemSkill improves the success rate from 75.37 to 83.58 while reducing the average number of steps from 20.69 to 16.63.

The advantage remains under transfer evaluation with Qwen, where MemSkill is not trained using this base model or dataset. MemSkill achieves an average success rate of 81.29, surpassing the strongest baseline Expel by 7.61 points. It also requires fewer interaction steps on both ALF-Seen and ALF-Unseen. These results suggest that the learned memory skills can capture reusable experience patterns for embodied decision-making, leading to both higher task success and more efficient execution across seen and unseen environments.

\begin{table*}[h]
    \footnotesize
    \centering
    \setlength{\tabcolsep}{5.5pt}
    \caption{\textbf{More comparison results on ALFWorld.}}
    \label{tab:more_exp_comparison}

    \begin{tabular}{>{\centering\arraybackslash}m{0.5cm} l cc cc c}
        \toprule
        \multirow{3}{*}{\textbf{Model}} &
        \multirow{3}{*}{\textbf{Methods}} &
        \multicolumn{5}{c}{\textbf{Embodied Interactive Tasks}} \\
        \cmidrule(lr){3-7}
        & &
        \multicolumn{2}{c}{\textbf{ALF-Seen$^\dagger$}} &
        \multicolumn{2}{c}{\textbf{ALF-Unseen$^\dagger$}} &
        \multicolumn{1}{c}{\textbf{Avg.}} \\
        \cmidrule(lr){3-4} \cmidrule(lr){5-6} \cmidrule(lr){7-7}
        & &
        \textbf{SR} & \textbf{\#Stps$\downarrow$} &
        \textbf{SR} & \textbf{\#Stps$\downarrow$} &
        \textbf{SR} \\
        \midrule

        % ===================== Base model block 1 =====================
        \multirow{4}{*}{\rotatebox[origin=c]{90}{\shortstack{\textbf{LLaMA}}}} &
        LightMem & 74.29 & 21.69 & 75.37 & 20.69 & 74.83 \\
        & AWM   & 66.43 & 23.25 & 68.66 & 22.25 & 67.55 \\
        & Expel   & 67.14 & 23.26 & 66.42 & 22.40 & 66.78 \\
        & \cellcolor{hl-llama}\textbf{MemSkill} &
        \cellcolor{hl-llama}\textbf{77.14} & \cellcolor{hl-llama}\textbf{18.91} &
        \cellcolor{hl-llama}\textbf{83.58} & \cellcolor{hl-llama}\textbf{16.63} &
        \cellcolor{hl-llama}\textbf{80.36} \\
        \midrule

        % ===================== Base model block 2 =====================
        \multirow{4}{*}{\rotatebox[origin=c]{90}{\shortstack{$^\blacktriangle$\textbf{Qwen}}}} &
        LightMem & 70.71 & 20.48 & 58.21 & 25.62 & 64.46 \\
        & AWM   & 74.29 & 19.03 & 61.19 & 24.67 & 67.74 \\
        & Expel   & 75.71 & 18.81 & 71.64 & 20.98 & 73.68 \\
        & \cellcolor{hl-qwen}\textbf{MemSkill} &
        \cellcolor{hl-qwen}\textbf{85.71} & \cellcolor{hl-qwen}\textbf{13.84} &
        \cellcolor{hl-qwen}\textbf{76.87} & \cellcolor{hl-qwen}\textbf{18.16} &
        \cellcolor{hl-qwen}\textbf{81.29} \\
        \bottomrule

        \multicolumn{7}{l}{\small \textbf{Bold} indicates the best score within each base model block.} \\
        \multicolumn{7}{l}{\small $^\blacktriangle$ indicates no training using this base model or dataset.} \\
        \multicolumn{7}{l}{\small $^\dagger$ indicates evaluation with in-context demonstrations.} \\
    \end{tabular}
\end{table*}

\begin{table*}[h]
    \footnotesize
    \centering
    \setlength{\tabcolsep}{5.5pt}
    \caption{\textbf{More comparison results on ALFWorld \textit{without} in-context demonstrations.}}
    \label{tab:more_exp_alfworld}

    \begin{tabular}{>{\centering\arraybackslash}m{0.5cm} l cc cc c}
        \toprule
        \multirow{3}{*}{\textbf{Model}} &
        \multirow{3}{*}{\textbf{Methods}} &
        \multicolumn{5}{c}{\textbf{Embodied Interactive Tasks}} \\
        \cmidrule(lr){3-7}
        & &
        \multicolumn{2}{c}{\textbf{ALF-Seen}} &
        \multicolumn{2}{c}{\textbf{ALF-Unseen}} &
        \multicolumn{1}{c}{\textbf{Avg.}} \\
        \cmidrule(lr){3-4} \cmidrule(lr){5-6} \cmidrule(lr){7-7}
        & &
        \textbf{SR} & \textbf{\#Stps$\downarrow$} &
        \textbf{SR} & \textbf{\#Stps$\downarrow$} &
        \textbf{SR} \\
        \midrule

        % ===================== Base model block 1 =====================
        \multirow{12}{*}{\rotatebox[origin=c]{90}{\shortstack{\textbf{LLaMA3.3}\\\textbf{70B-Instruct}}}} &
        No-Memory   & 17.14 & 43.74 & 20.15 & 42.99 & 18.65 \\
        & CoN   & 40.71 & 33.44 & 30.60 & 37.66 & 35.66 \\
        & ReadAgent   & 32.86 & 37.09 & 38.06 & 34.78 & 35.46 \\
        & MemoryBank   & 25.00 & 39.96 & 32.84 & 36.54 & 28.92 \\
        & A-MEM   & 24.29 & 40.51 & 28.36 & 38.83 & 26.33 \\
        & Mem0   & 32.86 & 36.47 & 32.09 & 37.32 & 32.48 \\
        & LangMem   & 37.86 & 34.39 & 35.07 & 35.70 & 36.47 \\
        & MemoryOS   & 15.71 & 43.74 & 14.18 & 44.54 & 14.95 \\
        & LightMem   & 45.71 & 31.76 & 46.27 & 30.66 & 45.99 \\
        & AWM   & 35.71 & 35.87 & 39.55 & 34.93 & 37.63 \\
        & Expel   & 42.14 & 34.61 & 43.28 & 33.25 & 42.71 \\
        & \cellcolor{hl-llama}\textbf{MemSkill} &
        \cellcolor{hl-llama}\textbf{55.71} & \cellcolor{hl-llama}\textbf{26.87} &
        \cellcolor{hl-llama}\textbf{59.70} & \cellcolor{hl-llama}\textbf{25.88} &
        \cellcolor{hl-llama}\textbf{57.71} \\
        \midrule

        % ===================== Base model block 2 =====================
        \multirow{12}{*}{\rotatebox[origin=c]{90}{\shortstack{$^\blacktriangle$\textbf{Qwen3-Next}\\\textbf{80B-A3B-Instruct}}}} &
        No-Memory   & 18.57 & 42.48 & 26.12 & 39.35 & 22.35 \\
        & CoN   & 57.86 & 25.81 & 53.73 & 28.40 & 55.80 \\
        & ReadAgent   & 53.57 & 27.88 & 54.48 & 27.41 & 54.03 \\
        & MemoryBank   & 37.86 & 35.15 & 38.06 & 34.99 & 37.96 \\
        & A-MEM   & 25.00 & 40.28 & 29.10 & 39.04 & 27.05 \\
        & Mem0   & 38.57 & 33.64 & 41.04 & 33.16 & 39.81 \\
        & LangMem   & 37.14 & 34.42 & 31.34 & 37.17 & 34.24 \\
        & MemoryOS   & 19.29 & 42.43 & 18.66 & 42.95 & 18.98 \\
        & LightMem   & 30.71 & 36.81 & 26.87 & 38.96 & 28.79 \\
        & AWM   & 47.86 & 31.24 & 47.01 & 31.69 & 47.44 \\
        & Expel   & 54.29 & 28.90 & 55.22 & 29.57 & 54.76 \\
        & \cellcolor{hl-qwen}\textbf{MemSkill} &
        \cellcolor{hl-qwen}\textbf{65.71} & \cellcolor{hl-qwen}\textbf{22.49} &
        \cellcolor{hl-qwen}\textbf{61.94} & \cellcolor{hl-qwen}\textbf{24.59} &
        \cellcolor{hl-qwen}\textbf{63.83} \\
        \bottomrule

        \multicolumn{7}{l}{\small \textbf{Bold} indicates the best score within each base model block.} \\
        \multicolumn{7}{l}{\small $^\blacktriangle$ indicates no training using this base model or dataset.} \\
    \end{tabular}
\end{table*}

Table~\ref{tab:more_exp_alfworld} further reports ALFWorld results \textit{without} in-context demonstrations. This setting is complementary to the main-table evaluation with demonstrations, but is more controlled for studying memory itself. Since in-context demonstrations can also act as an external form of memory, including them may confound the gains brought by each method's constructed memory. Therefore, this setting decouples demonstrations from memory construction and isolates the contribution of the learned memory mechanism.

Under this stricter setting, MemSkill still consistently outperforms all baselines across both base models and both evaluation splits. With LLaMA, MemSkill achieves an average success rate of 57.71, outperforming the strongest baseline LightMem by 11.72 points. The improvement is larger on ALF-Unseen, where MemSkill improves the success rate from 46.27 to 59.70, suggesting stronger generalization to unseen environments. MemSkill also requires fewer interaction steps than all baselines, reducing the average steps on ALF-Seen and ALF-Unseen to 26.87 and 25.88, respectively.

The same trend holds in the transfer setting with Qwen, where MemSkill is not trained using this base model or dataset. MemSkill achieves an average success rate of 63.83, improving over the strongest baseline by 8.03 points. It also obtains the lowest number of steps on both ALF-Seen and ALF-Unseen. These results show that the gains of MemSkill do not rely on in-context demonstrations, and that the learned memory skills provide reusable task experience that improves both success rate and execution efficiency.

\subsection{More Results on Appworld}
\label{appendix:more_exp_appworld}

\begin{table*}[t]
    \footnotesize
    \centering
    \setlength{\tabcolsep}{5pt}
    \caption{\textbf{More comparison results on AppWorld \textit{with} and \textit{without} in-context demonstrations.}}
    \label{tab:more_exp_appworld}

    \begin{tabular}{>{\centering\arraybackslash}m{0.5cm} l cc cc c | cc cc c}
        \toprule
        \multirow{3}{*}{\textbf{Model}} &
        \multirow{3}{*}{\textbf{Methods}} &
        \multicolumn{10}{c}{\textbf{Interactive Tool-Use Benchmark}} \\
        \cmidrule(lr){3-7} \cmidrule(lr){8-12}
        & &
        \multicolumn{2}{c}{\textbf{Test-N}} &
        \multicolumn{2}{c}{\textbf{Test-C}} &
        \multicolumn{1}{c|}{\textbf{Avg.}} &
        \multicolumn{2}{c}{\textbf{Test-N$^\dagger$}} &
        \multicolumn{2}{c}{\textbf{Test-C$^\dagger$}} &
        \multicolumn{1}{c}{\textbf{Avg.}} \\
        \cmidrule(lr){3-4} \cmidrule(lr){5-6} \cmidrule(lr){7-7}
        \cmidrule(lr){8-9} \cmidrule(lr){10-11} \cmidrule(lr){12-12}
        & &
        \textbf{PR} & \textbf{\#Steps$\downarrow$} &
        \textbf{PR} & \textbf{\#Steps$\downarrow$} &
        \textbf{PR} &
        \textbf{PR} & \textbf{\#Steps$\downarrow$} &
        \textbf{PR} & \textbf{\#Steps$\downarrow$} &
        \textbf{PR} \\
        \midrule

        % ===================== Base model block 1 =====================
        \multirow{10}{*}{\rotatebox[origin=c]{90}{\shortstack{\textbf{LLaMA3.3}\\\textbf{70B-Instruct}}}} &
        No-Memory   & 22.80 & 36.71 & 19.10 & 37.92 & 20.95 & 54.03 & 14.21 & 38.78 & 19.12 & 46.41 \\
        & ReadAgent  & 23.43 & 34.82 & 20.49 & 31.36 & 21.96 & 53.76 & 14.54 & 38.67 & 19.24 & 46.22 \\
        & MemoryBank & 23.28 & 35.66 & 19.96 & 33.34 & 21.62 & 53.66 & 14.89 & 38.46 & 18.84 & 46.06 \\
        & A-MEM      & 22.13 & 38.46 & 17.60 & 38.46 & 19.87 & 52.85 & 14.31 & 38.16 & 18.78 & 45.51 \\
        & Mem0       & 27.02 & 27.04 & 21.62 & 30.64 & 24.32 & 54.32 & 14.14 & 39.73 & 18.55 & 47.03 \\
        & MemoryOS   & 23.24 & 38.26 & 19.17 & 36.17 & 21.21 & 53.14 & 13.62 & 38.52 & 17.83 & 45.83 \\
        & LightMem   & 25.42 & 30.73 & 21.32 & 30.33 & 23.37 & 51.25 & 14.10 & 39.12 & 18.50 & 45.19 \\
        & AWM        & 29.00 & \textbf{17.73} & 21.83 & \textbf{20.48} & 25.42 & 54.68 & 14.26 & 39.54 & 18.64 & 47.11 \\
        & Expel      & 25.29 & 29.96 & 21.20 & 30.63 & 23.25 & 52.88 & 13.56 & 39.49 & 17.91 & 46.19 \\
        & \cellcolor{hl-llama}\textbf{MemSkill} &
        \cellcolor{hl-llama}\textbf{31.12} & \cellcolor{hl-llama}29.38 &
        \cellcolor{hl-llama}\textbf{22.29} & \cellcolor{hl-llama}32.17 &
        \cellcolor{hl-llama}\textbf{26.71} &
        \cellcolor{hl-llama}\textbf{54.84} & \cellcolor{hl-llama}\textbf{13.25} &
        \cellcolor{hl-llama}\textbf{39.79} & \cellcolor{hl-llama}\textbf{17.59} &
        \cellcolor{hl-llama}\textbf{47.32} \\
        \midrule

        % ===================== Base model block 2 =====================
        \multirow{10}{*}{\rotatebox[origin=c]{90}{\shortstack{$^\blacktriangle$\textbf{Qwen3-Next}\\\textbf{80B-A3B-Instruct}}}} &
        No-Memory   & 22.41 & 38.39 & 19.99 & 36.78 & 21.20 & 47.68 & 16.81 & 38.42 & 19.70 & 43.05 \\
        & ReadAgent  & 26.29 & 34.77 & 20.36 & 35.07 & 23.33 & 46.83 & 17.59 & 37.65 & 21.12 & 42.24 \\
        & MemoryBank & 24.41 & 36.94 & 20.28 & 36.19 & 22.35 & 46.50 & 16.98 & 37.18 & 22.03 & 41.84 \\
        & A-MEM      & 23.03 & 37.17 & 18.39 & 37.43 & 20.71 & 45.44 & 17.49 & 38.14 & 21.40 & 41.79 \\
        & Mem0       & 40.42 & \textbf{18.48} & 22.16 & 22.97 & 31.29 & 50.19 & 15.11 & 39.64 & 20.33 & 44.92 \\
        & MemoryOS   & 25.00 & 35.98 & 21.46 & 35.20 & 23.23 & 50.98 & 16.11 & 38.03 & 21.32 & 44.51 \\
        & LightMem   & 36.90 & 23.71 & 23.44 & 27.05 & 30.17 & 48.91 & 17.23 & 40.85 & 21.14 & 44.88 \\
        & AWM        & 27.02 & 23.90 & 22.05 & 22.63 & 24.54 & 53.23 & 16.77 & 41.26 & 22.95 & 47.25 \\
        & Expel      & 25.00 & 22.20 & 20.40 & \textbf{21.00} & 22.70 & 49.88 & 16.86 & 39.17 & 20.06 & 44.53 \\
        & \cellcolor{hl-qwen}\textbf{MemSkill} &
        \cellcolor{hl-qwen}\textbf{43.05} & \cellcolor{hl-qwen}26.11 &
        \cellcolor{hl-qwen}\textbf{25.41} & \cellcolor{hl-qwen}31.39 &
        \cellcolor{hl-qwen}\textbf{34.23} &
        \cellcolor{hl-qwen}\textbf{54.53} & \cellcolor{hl-qwen}\textbf{15.05} &
        \cellcolor{hl-qwen}\textbf{42.30} & \cellcolor{hl-qwen}\textbf{19.23} &
        \cellcolor{hl-qwen}\textbf{48.42} \\
        \bottomrule

        \multicolumn{12}{l}{\small \textbf{Bold} indicates the best score within each base model block.} \\
        \multicolumn{12}{l}{\small $^\blacktriangle$ indicates no training using this base model or dataset.} \\
        \multicolumn{12}{l}{\small $^\dagger$ indicates evaluation with in-context demonstrations.} \\
    \end{tabular}
\end{table*}

To further evaluate whether memory skills generalize beyond conversational and embodied household settings, we extend our study to AppWorld~\citep{trivedi2024appworld}, a more challenging interactive tool-use benchmark. We report results on both Test-Normal (Test-N) and Test-Challenge (Test-C), using Pass Rate (PR) and the average number of execution steps as evaluation metrics. Our default setting follows the stronger evaluation protocol with in-context demonstrations. In addition, we also report a controlled variant without demonstrations, which helps isolate the contribution of each method's constructed memory from the task guidance provided by demonstrations.

As shown in Table~\ref{tab:more_exp_appworld}, MemSkill achieves the best average PR across both base models under the default demonstration-based setting. With LLaMA, the No-Memory baseline already performs strongly, indicating that in-context demonstrations provide substantial task-specific guidance and make this setting partially saturated. As a result, the performance gaps among different memory methods become relatively small. Nevertheless, MemSkill still achieves the best overall pass rate and uses the fewest execution steps across both splits, suggesting that the learned memory skills can still improve efficiency even when demonstrations already provide strong external guidance.

The benefit of MemSkill is more evident with Qwen, where the demonstration-based setting leaves more room for memory-based improvement. In this case, MemSkill consistently outperforms prior memory methods and also improves over the No-Memory baseline. It further achieves the lowest execution steps on both Test-N and Test-C, indicating that MemSkill can effectively complement in-context demonstrations and support more efficient tool-use behavior.

The controlled setting without in-context demonstrations further confirms the independent contribution of learned memory skills. In this setting, MemSkill achieves the best average PR under both base models, with consistent gains on both Test-N and Test-C. This shows that MemSkill is not merely benefiting from demonstrations, but can provide useful experience abstraction when the agent must rely more directly on constructed memory.

Overall, these results show that MemSkill generalizes to interactive tool-use environments beyond conversational and embodied household tasks. Under the default demonstration-based setting, MemSkill remains competitive even when demonstrations provide strong task guidance, and brings clearer gains when the base model leaves more room for memory-based improvement. Under the controlled no-demonstration setting, MemSkill consistently improves pass rate over prior memory methods, further supporting the effectiveness of skill-conditioned memory construction for complex multi-step tool use.

\subsection{Experimental Results on Small Models}
\label{appendix:small_model}

We further evaluate MemSkill with Llama-3.1-8B-Instruct to examine whether the learned memory skills remain effective when the base model has more limited capacity. As shown in Table~\ref{tab:exp_small_model}, MemSkill consistently achieves the best performance across both LoCoMo and LongMemEval. Compared with prior memory methods, MemSkill improves both F1 and L-J on the two conversational benchmarks, indicating that skill-conditioned memory construction can provide useful support even for smaller backbone models.

The improvement is also maintained on LongMemEval, where no training is performed using this base model or dataset. This suggests that the learned memory skills are not tightly coupled to a specific backbone, and can transfer to smaller models under long-context conversational settings. Overall, these results show that MemSkill does not rely solely on the capability of a large base model. Instead, its learned memory skills provide complementary gains that remain effective under a more resource-efficient model setting.

\begin{table*}[h]
    \footnotesize
    \centering
    \setlength{\tabcolsep}{5.5pt}
    \caption{\textbf{Experimental results on LoCoMo and LongMemEval using Llama-3.1-8B-Instruct.}}
    \label{tab:exp_small_model}

    \begin{tabular}{>{\centering\arraybackslash}m{0.5cm} l cc cc c}
        \toprule
        \multirow{3}{*}{\textbf{Model}} &
        \multirow{3}{*}{\textbf{Methods}} &
        \multicolumn{5}{c}{\textbf{Conversational Benchmarks}} \\
        \cmidrule(lr){3-7}
        & &
        \multicolumn{2}{c}{\textbf{LoCoMo}} &
        \multicolumn{2}{c}{\textbf{$^\blacktriangle$LongMemEval}} &
        \multicolumn{1}{c}{\textbf{Avg.}} \\
        \cmidrule(lr){3-4} \cmidrule(lr){5-6} \cmidrule(lr){7-7}
        & &
        \textbf{F1} & \textbf{L-J} &
        \textbf{F1} & \textbf{L-J} &
        \textbf{L-J} \\
        \midrule

        % ===================== Base model block 1 =====================
        \multirow{8}{*}{\rotatebox[origin=c]{90}{\shortstack{\textbf{LLaMA3.1}\\\textbf{8B-Instruct}}}} &
        ReadAgent  & 17.53 & 21.49 & 9.45 & 18.80 & 20.15 \\
        & MemoryBank & 18.84 & 22.36 & 13.27 & 24.59 & 23.48 \\
        & A-MEM      & 19.95 & 25.62 & 16.12 & 25.71 & 25.67 \\
        & Mem0       & 14.26 & 17.12 & 14.59 & 26.38 & 21.75 \\
        & LangMem    & 13.36 & 15.71 & 6.52 & 17.16 & 16.44 \\
        & MemoryOS   & 17.68 & 23.08 & 7.88 & 18.24 & 20.66 \\
        & LightMem   & 19.74 & 26.85 & 17.68 & 29.40 & 28.13 \\
        & \cellcolor{hl-llama}\textbf{MemSkill} &
        \cellcolor{hl-llama}\textbf{20.59} & \cellcolor{hl-llama}\textbf{27.23} & \cellcolor{hl-llama}\textbf{18.13} & \cellcolor{hl-llama}\textbf{30.20} &
        \cellcolor{hl-llama}\textbf{28.72} \\
        \bottomrule

        \multicolumn{7}{l}{\small \textbf{Bold} indicates the best score within each base model block.} \\
        \multicolumn{7}{l}{\small $^\blacktriangle$ indicates no training using this base model or dataset.}
    \end{tabular}
\end{table*}

\subsection{Training Stability}
\label{appendix:train_stability}

MemSkill incorporates several mechanisms to improve the stability of skill learning and evolution. First, we maintain a snapshot of the skill bank after each evolution round. If the updated skill bank underperforms the best-performing snapshot observed so far, we roll back to the best snapshot and continue subsequent evolution from it. This prevents occasional harmful skill updates from permanently degrading the memory system.

Second, hard cases are selected according to difficulty scores, so the designer is guided by recurring failure patterns rather than noisy random examples. This encourages each evolution round to focus on systematic weaknesses of the current skill bank. Third, skill evolution is performed at controlled intervals, with a capped number of skill modifications in each round. This avoids abrupt changes to the skill bank and makes the learning process more gradual.

Together, these mechanisms make the evolution process less sensitive to noisy feedback and reduce the risk of unstable skill drift. In practice, we observe that the learned skill bank improves progressively over evolution rounds, suggesting that the snapshot rollback, hard-case selection, and controlled update strategy provide a stable basis for self-evolving memory skills.
More details can be found in Appendix~\ref{appendix:detail_component}

\subsection{Additional Analysis on Skill Selection}
\label{app:skill_selection}

We further evaluate the learned controller against two alternative skill selection strategies on LoCoMo. \textbf{Frozen Similarity} selects Top-$K$ skills using cosine similarity between frozen state and skill embeddings. \textbf{LLM Selector} prompts the base LLM to select Top-$K$ skills given the current span, retrieved memories, and full skill index. Both baselines use the same skill bank and executor as MemSkill.

\begin{table*}[h]
\centering
\caption{\textbf{Comparison of skill selection strategies on LoCoMo using LLaMA.}}
\label{tab:skill_selection}
\begin{tabular}{lcc}
\toprule
Method & F1 & L-J \\
\midrule
Frozen Similarity & 40.97 & 50.80 \\
LLM Selector & 39.16 & 48.73 \\
\cellcolor{hl-llama}\textbf{MemSkill} & \cellcolor{hl-llama}\textbf{44.21} & \cellcolor{hl-llama}\textbf{53.82} \\
\bottomrule
\end{tabular}
\end{table*}

As shown in Table~\ref{tab:skill_selection}, MemSkill consistently outperforms both alternatives, demonstrating that reward-optimized skill selection is more effective than static semantic similarity or direct LLM prompting.

\subsection{Comparison with Simplified Alternatives}
\label{app:simplified_alternatives}

We further compare MemSkill with two simplified alternatives on LoCoMo using LLaMA.

\textbf{Gradient-Free Self-Correction.}
We replace the learned controller with a prompted LLM selector throughout training and inference. The Designer continues to evolve the skill bank from downstream failure cases, but no policy optimization is performed.

\textbf{Static Well-Engineered Skill Set.}
Starting from MemSkill's best-performing skill bank, we further optimize its instructions using GPT-5.6-Sol-Pro, guided by existing memory systems and benchmark feedback, followed by manual verification. The resulting skill bank is then frozen and evaluated with a prompted LLM selector, without PPO training or further evolution.

\begin{table*}[h]
\centering
\caption{\textbf{Comparison with simplified alternatives on LoCoMo using LLaMA.}}
\label{tab:simplified_alternatives}
\begin{tabular}{lcc}
\toprule
Method & F1 & L-J \\
\midrule
Gradient-Free Self-Correction & 36.35 & 46.34 \\
Static Well-Engineered Skills & 35.28 & 44.90 \\
\cellcolor{hl-llama}\textbf{MemSkill} & \cellcolor{hl-llama}\textbf{44.21} & \cellcolor{hl-llama}\textbf{53.82} \\
\bottomrule
\end{tabular}
\end{table*}

As shown in Table~\ref{tab:simplified_alternatives}, MemSkill outperforms both alternatives, demonstrating the benefits of combining learned skill selection with feedback-driven skill evolution over purely prompted or static skill management.

\section{More Implementation Details}
\label{appendix:imple_detail}

\subsection{Evaluation Details}
\label{appendix:exp_eval}

\textbf{LLM judge and infrastructure.}
We use \texttt{openai/gpt-oss-120b} as the LLM judge (judge prompts can be found in Appendix~\ref{appendix:prompts}). All API-based models are accessed through the NV NIM API\footnote{\url{https://docs.nvidia.com/nim/}} and Together API\footnote{\url{https://docs.together.ai/}}. Training is conducted on NVIDIA A6000 GPUs.

\textbf{Training cost.}
On LoCoMo, a typical training run takes approximately 8--12 hours to converge in our setup, using a single NVIDIA A6000 GPU and consuming on the order of hundreds of millions of input tokens and tens of millions of output tokens. Training is primarily constrained by remote LLM inference and API latency, as PPO only optimizes lightweight MLPs. Skill evolution typically involves 5--6 rounds before convergence, with each Designer update taking a few minutes. These estimates may vary with API conditions and training configurations.

\textbf{LoCoMo~\citep{maharana2024evaluating-locomo}.}
LoCoMo contains 10 long interaction samples, each paired with roughly 200 training queries on average. We split the dataset by sample into train, validation, and test sets with a 6/2/2 ratio. We further remove \emph{adversarial} queries, since their supporting evidence is absent from the provided context and may introduce noisy supervision during training.

\textbf{LongMemEval~\citep{wu2024longmemeval}.}
We use the LongMemEval-S split, where each example contains an ultra-long conversation of roughly 100K tokens. We remove abstention questions, since they do not require retrieving or constructing useful memory from the conversation history. We split the remaining data into train, validation, and test sets. Although the training split is not used for learning in this transfer setting, the validation split is used to tune dataset-specific configurations. We then conduct transfer evaluation on a stratified test sample of about one-fifth of the dataset, approximately 100 samples, ensuring coverage of different question types for a comprehensive assessment.

\textbf{ALFWorld~\citep{shridhar2020alfworld}.}
We first collect expert trajectories from the training split and use them as the corpus for memory or experience construction. We then evaluate on the official ALF-Seen and ALF-Unseen splits. More training configuration details can be found in Appendix~\ref{appendix:detail_alfworld}.

\textbf{HotpotQA~\citep{yang2018hotpotqa}.}
We use HotpotQA to study transfer under distribution shift, following the evaluation protocol of~\citep{yu2025memagent}. Specifically, we evaluate on three context-length settings with increasing difficulty, corresponding to 50, 100, and 200 concatenated documents, denoted as \texttt{eval\_50}, \texttt{eval\_100}, and \texttt{eval\_200}. Unless otherwise specified, all results in this part use LLaMA as the base model and report the LLM-judge score (L-J).

\textbf{Span-level evaluation.}
During evaluation, we construct memory at the span level with a default span size of 512 tokens, rather than updating memory turn by turn. This substantially reduces the number of LLM calls and improves evaluation efficiency.

\subsection{More Details of the Designer}
\label{appendix:detail_component}

\paragraph{Hard-case buffer and representative case mining.}
The designer maintains a sliding \emph{hard-case buffer} that tracks recently challenging evaluation cases without growing unbounded. 
Each case stores the query, the retrieved memories used to answer it, the model prediction, the reference answer, the resulting task reward (e.g., F1), and a failure counter that records how many times the case has been answered incorrectly.
To prioritize cases that are both low-reward and repeatedly failed, we assign each case a difficulty score
\begin{equation}
d(q) \;=\; \big(1 - r(q)\big)\cdot c(q),
\end{equation}
where $r(q)\in[0,1]$ is the task reward for query $q$ and $c(q)$ is its cumulative failure count within the buffer window. Higher $d(q)$ indicates more critical cases that should be examined first.

To encourage coverage over \emph{diverse} failure types, we further cluster hard cases by semantic similarity of their queries and mine representative cases from each cluster.
For example, in LoCoMo, some queries focus on temporal cues (e.g., \emph{when} an event happened) while others emphasize locations (e.g., \emph{where} something occurred). 
Clustering helps separate these semantic types so the designer feedback is not dominated by a single frequent error mode, improving diversity and completeness of the mined supervision.

\paragraph{Exploration incentive for newly introduced skills.}
After each evolution round, the designer may introduce new skills that the controller has not yet learned to utilize.
To facilitate adoption, we apply a short post-update exploration phase by biasing the controller toward new skills directly at the logit level.
Let $\mathcal{S}_{\text{new}} \subseteq \mathcal{S}_t$ denote the newly added skills, and let $p_\theta(i\mid h_t)=\mathrm{softmax}(z_t)_i$ be the controller distribution at step $t$.
We encourage the total probability mass assigned to new skills to reach a target threshold $\tau_q$:
\begin{equation}
\sum_{i\in \mathcal{S}_{\text{new}}} p_\theta(i\mid h_t)\;\ge\;\tau_q,
\qquad \tau_q\in[0,1].
\label{eq:new_skill_mass_constraint}
\end{equation}
When the constraint in Eq.~\eqref{eq:new_skill_mass_constraint} is violated, we add a uniform logit gain $\delta_q$ to all new skills,
\begin{equation}
z'_{t,i} \;=\;
\begin{cases}
z_{t,i} + \delta_q, & i\in \mathcal{S}_{\text{new}},\\
z_{t,i}, & \text{otherwise},
\end{cases}
\qquad
p'_\theta(\cdot\mid h_t)=\mathrm{softmax}(z'_t),
\end{equation}
where $\delta_q$ is chosen as the minimal value that makes $\sum_{i\in \mathcal{S}_{\text{new}}} p'_\theta(i\mid h_t)\ge\tau_q$.
By operating on logits, this mechanism preserves the controller architecture and provides a smooth probability-level encouragement toward new skills.

We apply this incentive for the first $T_{\text{explore}}{=}50$ training steps after each evolution round.
To avoid persistent bias, the target threshold decays linearly within this window:
\begin{equation}
\tau_q \;=\; \tau_0 \cdot \Big(1 - \frac{q}{T_{\text{explore}}}\Big),
\qquad q=0,1,\dots,T_{\text{explore}}-1,
\end{equation}
with default $\tau_0{=}0.3$.
This schedule provides strong initial exploration and then gradually fades, yielding a smooth transition back to the controller's learned selection behavior.

\paragraph{Early stopping and rollback based on stabilized rewards.}
MemSkill performs skill evolution periodically, where each evolution cycle consists of a fixed number of controller-training steps (e.g., 100 steps) on the current skill bank. Because the reward signal can be volatile immediately after a skill-bank update, we assess whether a cycle improves performance using a \emph{stabilized} reward estimate: we compute the average task reward over the \emph{last quarter} of training steps within the cycle, and treat this value as the cycle's score.

Let $L$ denote the number of controller-training steps per cycle and $\{r_t\}_{t=1}^{L}$ the step-level rewards within the cycle. We define the cycle score as
\begin{equation}
\bar{r}_{\text{tail}} \;=\; \frac{1}{L/4}\sum_{t=3L/4+1}^{L} r_t .
\end{equation}
We compare $\bar{r}_{\text{tail}}$ against the best score observed so far. If the current cycle does not improve this criterion, then before performing the next skill evolution step, we roll back the skill bank to the previously best-performing snapshot and restart evolution from that snapshot. This rollback prevents compounding degradations from suboptimal designer updates.

Finally, if the stabilized reward fails to improve for several consecutive evolution cycles (we use a fixed patience), we early stop training and return the best skill bank snapshot encountered during training.

\subsection{Details on ALFWorld Training}
\label{appendix:detail_alfworld}

ALFWorld differs from the other benchmarks in that it is an interactive environment rather than a static text corpus. To instantiate MemSkill in this setting, we first convert ALFWorld into an offline training protocol by collecting expert trajectories on the training split. Each trajectory records the agent's interaction sequence (observations, actions, and outcomes) and serves as an interaction trace for memory construction.

\paragraph{Task-type grouping.}
ALFWorld tasks naturally fall into a small number of recurring goal templates. Following common practice, we group trajectories by task type (i.e., goal template), such as \textsc{Pick \& Place} (put an object into/on a target receptacle), \textsc{Clean \& Place} (clean an object and then place it), \textsc{Heat \& Place} (heat an object and then place it), and \textsc{Cool \& Place} (cool an object and then place it).\footnote{We use the task template provided by the environment to define task types.}

\paragraph{Experience corpus vs.\ evaluation cases.}
To fit ALFWorld into our training framework, we construct per-type train-time data splits from the offline expert trajectories. For each task type, we randomly sample a subset of trajectories as the \emph{experience corpus} used for memory construction, and sample another \emph{non-overlapping} subset of trajectories from the same type as \emph{evaluation cases}. During training, MemSkill builds a trajectory-specific memory bank from the experience corpus (span by span, via controller and executor), and then evaluates the constructed memory on the evaluation cases to obtain task reward and to log failure cases.

\paragraph{Motivation.}
Using non-overlapping trajectories from the \emph{same} task type for experience construction and evaluation provides a controlled generalization signal: trajectories within a type share goal structure and recurrent interaction patterns, making memories and skills more transferable across different instances of the same template. This setup encourages MemSkill to learn reusable memory skills that capture type-level regularities (e.g., relevant object states and action prerequisites) rather than overfitting to a single trajectory, while still ensuring that evaluation traces are held out from the traces used to build memory.

\subsection{Details on Training Objectives}
\label{appendix:rl_objective}

This part details the reinforcement learning objective used to optimize the controller in MemSkill when each decision selects an ordered Top-$K$ \emph{set} of skills without replacement.

\paragraph{Episode, states, and Top-$K$ actions.}
Training iterates over interaction traces (episodes). For a trace, MemSkill processes spans sequentially. 
At step $t$, the controller observes the raw state $s_t \triangleq (x_t,M_t)$, represented by the learned state embedding $h_t$ defined in Section~\ref{sec:controller}.
Let $\mathcal{S}_t=\{1,\dots,N_t\}$ denote the current skill bank, whose size $N_t$ may change as the designer evolves skills.
The controller computes logits $z_t\in\mathbb{R}^{N_t}$ by applying the shared scorer to all state-skill pairs, and induces
\begin{equation}
p_\theta(i\mid h_t)=\mathrm{softmax}(z_t)_i.
\end{equation}
Instead of sampling a single skill, the controller selects an \emph{ordered} Top-$K$ set
$A_t=(a_{t,1},\dots,a_{t,K})$ \emph{without replacement}, implemented via Gumbel-Top-$K$ sampling~\citep{kool2019stochastic} (i.e., adding i.i.d.\ Gumbel noise to logits and taking the top-$K$ indices).

\paragraph{Joint probability of Top-$K$ without-replacement selection.}
For PPO-style policy optimization, we need the joint probability of sampling the ordered set $A_t$ under the without-replacement process. This probability can be written as
\begin{equation}
\pi_\theta(A_t \mid h_t)
= \prod_{j=1}^{K}
\frac{p_\theta(a_{t,j}\mid h_t)}
{1-\sum_{\ell<j} p_\theta(a_{t,\ell}\mid h_t)} ,
\label{eq:topk_joint_prob_appendix}
\end{equation}
with the corresponding joint log-probability
\begin{equation}
\log \pi_\theta(A_t \mid h_t)
= \sum_{j=1}^{K}\Big(
\log p_\theta(a_{t,j}\mid h_t)
-\log\big(1-\sum_{\ell<j} p_\theta(a_{t,\ell}\mid h_t)\big)
\Big).
\label{eq:topk_joint_logprob_appendix}
\end{equation}
When $K=1$, Eq.~\ref{eq:topk_joint_prob_appendix} reduces to the standard single-action case.

\paragraph{Rewards from memory-dependent evaluation.}
For each trace, after processing all spans and constructing the trace-specific memory bank, we evaluate the memory bank on the trace's memory-dependent training queries and obtain a scalar task score (e.g., F1 or success rate). We treat this score as the episode-level reward:
\begin{equation}
R \triangleq \mathrm{Eval}(\text{memory bank}; \text{training queries}) \in \mathbb{R}.
\end{equation}
This reward is then assigned to the sequence of controller decisions within the trace. Concretely, we use standard return computation with discount factor $\gamma$:
\begin{equation}
G_t=\sum_{\tau=t}^{T}\gamma^{\tau-t}r_\tau,
\end{equation}
where $r_\tau$ is the per-step reward. In our default setting, reward is provided only after memory construction completes, i.e., $r_T=R$ and $r_\tau=0$ for $\tau<T$, so $G_t=\gamma^{T-t}R$.
We learn a value function $V_\phi(h_t)$ and compute advantages $\hat{A}_t$ using generalized advantage estimation (GAE).

\paragraph{PPO objective with Top-$K$ actions.}
We optimize the controller using proximal policy optimization (PPO)~\citep{schulman2017proximal}, replacing the standard single-action log-probability with the Top-$K$ joint log-probability in Eq.~\ref{eq:topk_joint_logprob_appendix}.
Let $\theta_{\text{old}}$ denote the parameters of the behavior policy used to collect rollouts.
For simplicity, we use $h_t$ to denote the state representation computed under the policy being evaluated.
Define the importance ratio:
\begin{equation}
r_t(\theta)=
\frac{\pi_\theta(A_t\mid h_t)}{\pi_{\theta_{\text{old}}}(A_t\mid h_t)}
=
\exp\Big(\log\pi_\theta(A_t\mid h_t)-\log\pi_{\theta_{\text{old}}}(A_t\mid h_t)\Big).
\label{eq:topk_ratio}
\end{equation}
The clipped surrogate policy objective is
\begin{equation}
\mathcal{L}_{\text{policy}}(\theta)
=
\mathbb{E}_t\Big[
\min\big(r_t(\theta)\,\hat{A}_t,\;
\mathrm{clip}(r_t(\theta),1-\epsilon,1+\epsilon)\,\hat{A}_t\big)
\Big].
\label{eq:ppo_policy_topk}
\end{equation}
We additionally optimize a value function and include an entropy bonus for exploration:
\begin{align}
\mathcal{L}_{\text{value}}(\phi)
&=
\mathbb{E}_t\Big[
\big(V_\phi(h_t)-G_t\big)^2
\Big], \label{eq:ppo_value}\\
\mathcal{H}(\theta)
&=
\mathbb{E}_t\big[H(p_\theta(\cdot\mid h_t))\big], \label{eq:ppo_entropy}
\end{align}
where $H(\cdot)$ is the entropy of the categorical distribution over all skills.
The overall objective (to maximize) is
\begin{equation}
\max_{\theta,\phi}\;
\mathcal{L}_{\text{policy}}(\theta)
- c_v\,\mathcal{L}_{\text{value}}(\phi)
+ c_H\,\mathcal{H}(\theta).
\label{eq:ppo_full_topk}
\end{equation}
In implementation, we minimize the negative of Eq.~\ref{eq:ppo_full_topk}.

\paragraph{Gumbel-Top-$K$ exploration.}
To sample Top-$K$ skills without replacement during rollout collection, we use Gumbel-Top-$K$ sampling: at each step we draw i.i.d.\ Gumbel noise $\{g_i\}_{i=1}^{N_t}$, form perturbed logits $\tilde{z}_{t,i}=z_{t,i}+g_i$, and take the indices of the $K$ largest $\tilde{z}_{t,i}$ to obtain $A_t$.
This provides stochastic exploration over skill subsets while remaining compatible with PPO through the joint probability in Eq.~\ref{eq:topk_joint_prob_appendix}.
For training stability, entropy regularization is computed from the base categorical distribution $p_\theta(\cdot\mid h_t)$ over all skills (Eq.~\ref{eq:ppo_entropy}), which encourages exploration of the evolving skill bank even though the executed action is a Top-$K$ set.

\section{Case Study}
\label{appendix:case_study}

\subsection{Initial Primitive Skills}

\begin{center}
\begin{tcolorbox}[title={Initial Primitive Skill - INSERT}]
{
Skill: Insert New Memory \\\\
Description: Memory management skill for capturing new, durable facts from the current text chunk that are not already in memory. \\\\
Purpose: Capture new, durable facts from the current text chunk that are missing in memory. \\\\
When to use: \\
- The text chunk introduces new facts, events, plans, or context worth storing. \\
- The information is stable and likely useful later. \\\\
How to apply: \\
- Compare against retrieved memories to avoid duplicates. \\
- Split distinct facts into separate items. \\
- Keep each item concise and specific. \\\\
Constraints: \\
- Skip trivial, fleeting, or speculative content. \\
- Do not update or delete existing memories. \\\\
Action type: INSERT only.
}
\label{appendix:case_insert}
\end{tcolorbox}
\end{center}

\begin{center}
\begin{tcolorbox}[title={Initial Primitive Skill - UPDATE}]
{
Skill: Update Existing Memory \\\\
Description: Memory management skill for revising an existing memory item when the text chunk provides corrections or new details. \\\\
Purpose: Revise a retrieved memory with new or corrected information from the text chunk. \\\\
When to use: \\
- The text chunk clarifies, corrects, or extends a retrieved memory. \\\\
How to apply: \\
- Select the best matching memory item. \\
- Merge new details into a single updated item. \\
- Preserve accurate details that still hold. \\\\
Constraints: \\
- Do not create new memories. \\
- Do not delete items. \\\\
Action type: UPDATE only.
}
\label{appendix:case_update}
\end{tcolorbox}
\end{center}

\begin{center}
\begin{tcolorbox}[title={Initial Primitive Skill - DELETE}]
{
Skill: Delete Invalid Memory \\\\
Description: Memory management skill for removing memory items that are incorrect, outdated, or superseded. \\\\
Purpose: Remove a retrieved memory that is wrong, outdated, or superseded by the text chunk. \\\\
When to use: \\
- The text chunk clearly contradicts a memory. \\
- A plan or fact is explicitly canceled or replaced. \\\\
How to apply: \\
- Only delete when evidence is explicit. \\\\
Constraints: \\
- If uncertain, prefer no action over deletion. \\\\
Action type: DELETE only.
}
\label{appendix:case_delete}
\end{tcolorbox}
\end{center}

\begin{center}
\begin{tcolorbox}[title={Initial Primitive Skill - SKIP}]
{
Skill: No Operation \\\\
Description: Memory management skill for confirming that no memory changes are required. \\\\
Purpose: Confirm no memory changes are needed for the text chunk. \\\\
When to use: \\
- The text chunk contains no new, corrective, or actionable information. \\\\
Constraints: \\
- Emit NOOP only if none of the selected skills produce actions. \\\\
Action type: NOOP only.
}
\label{appendix:case_noop}
\end{tcolorbox}
\end{center}

\subsection{Evolved Skills on LoCoMo}

% ---------- capture_activity_details ----------
\begin{center}
\begin{tcolorbox}[
  title={Evolved Skill on LoCoMo - CAPTURE\_ACTIVITY\_DETAILS},
  colback=blue!2!white,
  colframe=blue!35!black,
  colbacktitle=blue!15!white,
  coltitle=black,
  coltext=black
]
{
Skill: Capture Activity Details \\\\
Purpose: Capture detailed information about activities mentioned in the text chunk, including the type of activity, location, participants, temporal details, and any relevant contextual information. \\\\
When to use: \\
- The text chunk mentions a specific activity or event with contextual details. \\\\
How to apply: \\
- Identify the key elements of the activity (e.g., type, location, participants, temporal details). \\
- Capture any relevant contextual information that provides additional insight into the activity. \\
- Keep the activity details specific, actionable, and concise. \\\\
Constraints: \\
- Focus on explicit activity details and contextual information mentioned in the text chunk. \\
- Avoid inferring activity details or context not directly stated. \\\\
Action type: INSERT only.
}
\label{appendix:case_capture_activity_details}
\end{tcolorbox}
\end{center}

% ---------- capture_entity_nuances ----------
\begin{center}
\begin{tcolorbox}[
  title={Evolved Skill on LoCoMo - CAPTURE\_ENTITY\_NUANCES},
  colback=blue!2!white,
  colframe=blue!35!black,
  colbacktitle=blue!15!white,
  coltitle=black,
  coltext=black
]
{
Skill: Capture Entity Nuances \\\\
Purpose: Capture nuanced details about entities mentioned in the text chunk, such as nicknames, aliases, or comparative statements. \\\\
When to use: \\
- The text chunk mentions an entity with nuanced details (e.g., nickname, alias, comparison). \\\\
How to apply: \\
- Identify the entity and its associated nuances. \\
- Capture these nuances in a way that distinguishes them from the entity's primary information. \\\\
Constraints: \\
- Focus on explicit nuances mentioned in the text chunk. \\
- Avoid inferring nuances not directly stated. \\\\
Action type: INSERT only.
}
\label{appendix:case_capture_entity_nuances}
\end{tcolorbox}
\end{center}

% ---------- capture_temporal_context ----------
\begin{center}
\begin{tcolorbox}[
  title={Evolved Skill on LoCoMo - CAPTURE\_TEMPORAL\_CONTEXT},
  colback=blue!2!white,
  colframe=blue!35!black,
  colbacktitle=blue!15!white,
  coltitle=black,
  coltext=black
]
{
Skill: Capture Temporal Context \\\\
Purpose: Capture the temporal context of events, activities, or facts mentioned in the text chunk, including any relevant dates, times, durations, or sequential information. \\\\
When to use: \\
- The text chunk mentions a specific event, activity, or fact with associated temporal information. \\\\
How to apply: \\
- Identify the key temporal elements (e.g., start time, end time, duration, sequence). \\
- Capture the temporal context in a concise and specific format, considering any sequential relationships. \\\\
Constraints: \\
- Focus on explicit temporal information mentioned in the text chunk. \\
- Avoid inferring temporal details not directly stated. \\\\
Action type: INSERT only.
}
\label{appendix:case_capture_temporal_context}
\end{tcolorbox}
\end{center}

% ---------- delete ----------
\begin{center}
\begin{tcolorbox}[
  title={Evolved Skill on LoCoMo - DELETE},
  colback=blue!2!white,
  colframe=blue!35!black,
  colbacktitle=blue!15!white,
  coltitle=black,
  coltext=black
]
{
Skill: Delete Invalid Memory \\\\
Purpose: Remove a retrieved memory that is wrong, outdated, or superseded by the text chunk. \\\\
When to use: \\
- The text chunk clearly contradicts a memory. \\
- A plan or fact is explicitly canceled or replaced. \\\\
How to apply: \\
- Only delete when evidence is explicit. \\\\
Constraints: \\
- If uncertain, prefer no action over deletion. \\\\
Action type: DELETE only.
}
\label{appendix:case_delete}
\end{tcolorbox}
\end{center}

% ---------- handle_entity_relationships ----------
\begin{center}
\begin{tcolorbox}[
  title={Evolved Skill on LoCoMo - HANDLE\_ENTITY\_RELATIONSHIPS},
  colback=blue!2!white,
  colframe=blue!35!black,
  colbacktitle=blue!15!white,
  coltitle=black,
  coltext=black
]
{
Skill: Handle Entity Relationships \\\\
Purpose: Capture and manage complex relationships between entities mentioned in the text chunk, including nuanced details. \\\\
When to use: \\
- The text chunk mentions interactions, associations, or relationships between entities with specific details. \\\\
How to apply: \\
- Identify the entities involved and their roles in the relationship. \\
- Capture the nature of the relationship and any nuanced details (e.g., nicknames, comparative statements). \\
- Update existing memories to reflect the new relationship information. \\\\
Constraints: \\
- Focus on explicit relationships mentioned in the text chunk. \\
- Avoid inferring relationships not directly stated. \\\\
Action type: INSERT only.
}
\label{appendix:case_handle_entity_relationships}
\end{tcolorbox}
\end{center}

% ---------- insert ----------
\begin{center}
\begin{tcolorbox}[
  title={Evolved Skill on LoCoMo - INSERT},
  colback=blue!2!white,
  colframe=blue!35!black,
  colbacktitle=blue!15!white,
  coltitle=black,
  coltext=black
]
{
Skill: Insert New Memory \\\\
Purpose: Capture new, durable facts from the current text chunk that are missing in memory, including specific temporal details such as dates or time frames and detailed activity information. \\\\
When to use: \\
- The text chunk introduces new facts, events, plans, or context worth storing. \\
- The information is stable and likely useful later. \\\\
How to apply: \\
- Compare against retrieved memories to avoid duplicates. \\
- Split distinct facts into separate items. \\
- Keep each item concise and specific, including relevant temporal information and activity details. \\\\
Constraints: \\
- Skip trivial, fleeting, or speculative content. \\
- Do not update or delete existing memories. \\\\
Action type: INSERT only.
}
\label{appendix:case_insert}
\end{tcolorbox}
\end{center}

% ---------- noop ----------
\begin{center}
\begin{tcolorbox}[
  title={Evolved Skill on LoCoMo - NOOP},
  colback=blue!2!white,
  colframe=blue!35!black,
  colbacktitle=blue!15!white,
  coltitle=black,
  coltext=black
]
{
Skill: No Operation \\\\
Purpose: Confirm no memory changes are needed for the text chunk. \\\\
When to use: \\
- The text chunk contains no new, corrective, or actionable information. \\\\
Constraints: \\
- Emit NOOP only if none of the selected skills produce actions. \\\\
Action type: NOOP only.
}
\label{appendix:case_noop}
\end{tcolorbox}
\end{center}

% ---------- refine_temporal_details_with_context ----------
\begin{center}
\begin{tcolorbox}[
  title={Evolved Skill on LoCoMo - REFINE\_TEMPORAL\_DETAILS\_WITH\_CONTEXT},
  colback=blue!2!white,
  colframe=blue!35!black,
  colbacktitle=blue!15!white,
  coltitle=black,
  coltext=black
]
{
Skill: Refine Temporal Details with Context \\\\
Purpose: Update the temporal context of existing memories with new information from the text chunk, considering the context in which the information is provided. \\\\
When to use: \\
- The text chunk provides new or corrected temporal information relevant to an existing memory, and the context suggests a need for refinement. \\\\
How to apply: \\
- Identify the relevant existing memory and its current temporal context. \\
- Update the temporal details to reflect the new information, ensuring consistency with the provided context. \\\\
Constraints: \\
- Focus on explicit temporal information mentioned in the text chunk and supported by the context. \\
- Avoid inferring temporal details not directly stated or implied by the context. \\\\
Action type: UPDATE only.
}
\label{appendix:case_refine_temporal_details_with_context}
\end{tcolorbox}
\end{center}

% ---------- update ----------
\begin{center}
\begin{tcolorbox}[
  title={Evolved Skill on LoCoMo - UPDATE},
  colback=blue!2!white,
  colframe=blue!35!black,
  colbacktitle=blue!15!white,
  coltitle=black,
  coltext=black
]
{
Skill: Update Existing Memory \\\\
Purpose: Revise a retrieved memory with new or corrected information from the text chunk, including entity-specific details. \\\\
When to use: \\
- The text chunk clarifies, corrects, or extends a retrieved memory. \\
- The text chunk provides new information about a specific entity or its activities. \\\\
How to apply: \\
- Select the best matching memory item. \\
- Merge new details into a single updated item. \\
- Preserve accurate details that still hold, and ensure entity-specific information is accurately captured and updated. \\\\
Constraints: \\
- Do not create new memories. \\
- Do not delete items. \\\\
Action type: UPDATE only.
}
\label{appendix:case_update}
\end{tcolorbox}
\end{center}

\subsection{Evolved Skills on ALFWorld}

% ---------- capture_action_constraints ----------
\begin{center}
\begin{tcolorbox}[
  title={Evolved Skill on ALFWorld - CAPTURE\_ACTION\_CONSTRAINTS},
  colback=orange!6,
  colframe=orange!75!black,
  coltitle=black
]
{
Skill: Capture Action Constraints \\\\
Purpose: Capture detailed constraints on actions, including object states and movements, necessary for task completion. \\\\
When to use: \\
- The text chunk mentions constraints on actions, including object states and movements. \\
- The constraints are crucial for future task steps. \\\\
How to apply: \\
- Identify the action, its constraints, and relevant object states and movements from the text chunk. \\
- Create a new memory item with the action-constraint pair, including object states and movements. \\\\
Constraints: \\
- Only capture constraints on actions relevant to the task. \\
- Update existing constraint memories if new information is provided. \\\\
Action type: INSERT only.
}
\label{appendix:case_capture_action_constraints}
\end{tcolorbox}
\end{center}

% ---------- delete ----------
\begin{center}
\begin{tcolorbox}[
  title={Evolved Skill on ALFWorld - DELETE},
  colback=orange!6,
  colframe=orange!75!black,
  coltitle=black
]
{
Skill: Delete Invalid Memory \\\\
Purpose: Remove a retrieved memory that is wrong, outdated, or superseded by the text chunk. \\\\
When to use: \\
- The text chunk clearly contradicts a memory. \\
- A plan or fact is explicitly canceled or replaced. \\\\
How to apply: \\
- Only delete when evidence is explicit. \\\\
Constraints: \\
- If uncertain, prefer no action over deletion. \\\\
Action type: DELETE only.
}
\label{appendix:case_delete}
\end{tcolorbox}
\end{center}

% ---------- insert ----------
\begin{center}
\begin{tcolorbox}[
  title={Evolved Skill on ALFWorld - INSERT},
  colback=orange!6,
  colframe=orange!75!black,
  coltitle=black
]
{
Skill: Insert New Memory \\\\
Purpose: Capture new, durable facts, including procedural knowledge and action sequences, from the current text chunk that are missing in memory. \\\\
When to use: \\
- The text chunk introduces new facts, events, plans, or context worth storing. \\
- The information is stable and likely useful later. \\\\
How to apply: \\
- Compare against retrieved memories to avoid duplicates. \\
- Split distinct facts into separate items, including action sequences. \\
- Keep each item concise and specific. \\\\
Constraints: \\
- Skip trivial, fleeting, or speculative content. \\
- Do not update or delete existing memories. \\\\
Action type: INSERT only.
}
\label{appendix:case_insert}
\end{tcolorbox}
\end{center}

% ---------- noop ----------
\begin{center}
\begin{tcolorbox}[
  title={Evolved Skill on ALFWorld - NOOP},
  colback=orange!6,
  colframe=orange!75!black,
  coltitle=black
]
{
Skill: No Operation \\\\
Purpose: Confirm no memory changes are needed for the text chunk. \\\\
When to use: \\
- The text chunk contains no new, corrective, or actionable information. \\\\
Constraints: \\
- Emit NOOP only if none of the selected skills produce actions. \\\\
Action type: NOOP only.
}
\label{appendix:case_noop}
\end{tcolorbox}
\end{center}

% ---------- track_object_location ----------
\begin{center}
\begin{tcolorbox}[
  title={Evolved Skill on ALFWorld - TRACK\_OBJECT\_LOCATION},
  colback=orange!6,
  colframe=orange!75!black,
  coltitle=black
]
{
Skill: Track Object Location \\\\
Purpose: Explicitly track the location and state of an object necessary for task completion. \\\\
When to use: \\
- The text chunk mentions an object's location or state. \\
- The object's location or state is crucial for future task steps. \\\\
How to apply: \\
- Identify the object, its location, and relevant state from the text chunk. \\
- Create a new memory item with the object-location-state triplet. \\\\
Constraints: \\
- Only track locations and states of objects relevant to the task. \\
- Update existing location memories if new information is provided. \\\\
Action type: INSERT only.
}
\label{appendix:case_track_object_location}
\end{tcolorbox}
\end{center}

% ---------- track_object_movements ----------
\begin{center}
\begin{tcolorbox}[
  title={Evolved Skill on ALFWorld - TRACK\_OBJECT\_MOVEMENTS},
  colback=orange!6,
  colframe=orange!75!black,
  coltitle=black
]
{
Skill: Track Object Movements \\\\
Purpose: Track movements of objects necessary for task completion. \\\\
When to use: \\
- The text chunk mentions an object's movement. \\
- The object's movement is crucial for future task steps. \\\\
How to apply: \\
- Identify the object and its movement from the text chunk. \\
- Create a new memory item with the object-movement pair. \\\\
Constraints: \\
- Only track movements of objects relevant to the task. \\
- Update existing movement memories if new information is provided. \\\\
Action type: INSERT only.
}
\label{appendix:case_track_object_movements}
\end{tcolorbox}
\end{center}

% ---------- update ----------
\begin{center}
\begin{tcolorbox}[
  title={Evolved Skill on ALFWorld - UPDATE},
  colback=orange!6,
  colframe=orange!75!black,
  coltitle=black
]
{
Skill: Update Existing Memory \\\\
Purpose: Revise a retrieved memory with new or corrected information from the text chunk. \\\\
When to use: \\
- The text chunk clarifies, corrects, or extends a retrieved memory. \\\\
How to apply: \\
- Select the best matching memory item. \\
- Merge new details into a single updated item. \\
- Preserve accurate details that still hold. \\\\
Constraints: \\
- Do not create new memories. \\
- Do not delete items. \\\\
Action type: UPDATE only.
}
\label{appendix:case_update}
\end{tcolorbox}
\end{center}

\section{Prompts}
\label{appendix:prompts}

\begin{center}
\begin{tcolorbox}[title={LoCoMo Answer Prompt}]
{
Based on the above context, write an answer in the form of a short phrase for the following question. Answer with exact words from the context whenever possible. \\

Question: \{\} Short answer:
}
\label{appendix:case_locomo}
\end{tcolorbox}
\end{center}

\begin{center}
\begin{tcolorbox}[title={LongMemEval Answer Prompt}]
{
I will give you several history chats between you and a user. Please answer the question based on the relevant chat history. \\
History Chats: \{\} \\
Current Date: \{\} \\
Question: \{\} \\
Short Answer:
}
\label{appendix:case_longmemeval}
\end{tcolorbox}
\end{center}

\begin{center}
\begin{tcolorbox}[title={HotpotQA Answer Prompt}]
{
Based on the following context, answer the question. The question may require reasoning across multiple pieces of information. \\
\{context\}

Question: \{question\} \\

Instructions: \\
- Read the context carefully and identify relevant information \\
- If the answer can be found in the context, provide a short, precise answer \\
- Output your answer within \textless answer\textgreater \textless /answer\textgreater tags \\

\textless answer\textgreater your answer here\textless /answer\textgreater
}
\label{appendix:case_hp}
\end{tcolorbox}
\end{center}

\begin{center}
\begin{tcolorbox}[title={ALFWorld Env Interaction Prompt}]
{
You are controlling a text-based ALFWorld environment. Your job: choose the NEXT action as ONE text command. Output ONLY the command string, with no extra text. You MUST choose an action from the admissible actions list and copy it EXACTLY. \\\\
Goal: \{goal\} \\
Retrieved procedural tips (optional, short \& actionable): \{retrieved\_memory\} \\
Interaction history so far (most recent info matters most): \{history\} \\
Admissible actions (choose exactly ONE and copy it verbatim): \{actions\} \\\\
Now output exactly one line: the chosen action (must match one item above).
}
\label{appendix:case_alfworld}
\end{tcolorbox}
\end{center}

\begin{center}
\begin{tcolorbox}[title={Executor Prompt}]
{
You are a memory management executor. Apply the selected skills to the input text chunk and retrieved memories, then output memory actions. \\\\
Input Text Chunk: \{session\_text\} \\
Retrieved Memories (0-based index): \{mem\_text\} \\
Selected Skills: \{skills\_text\} \\\\
Guidelines: \\
- Apply any skill as needed; a skill may be used multiple times. \\
- Read the input text chunk carefully line by line and apply any skill as needed. \\
- Only use action types supported by the selected skills. \\
- MEMORY\_INDEX is 0-based and must reference the retrieved memories list. \\
- Output only action blocks in the format below. \\
- Do not include explanations or REASONING lines. \\\\
Output format (repeat as needed). Use ONE block per action and separate blocks with a blank line: \\\\
INSERT block: \\
ACTION: INSERT \\
MEMORY\_ITEM: [concise but complete summary with essential details] \\\\
UPDATE block: \\
ACTION: UPDATE \\
MEMORY\_INDEX: [0-based index] \\
UPDATED\_MEMORY: [concise but complete merged summary with essential updates] \\\\
DELETE block: \\
ACTION: DELETE \\
MEMORY\_INDEX: [0-based index]
}
\label{appendix:case_executor}
\end{tcolorbox}
\end{center}

\begin{center}
\begin{tcolorbox}[title={Designer Analysis Prompt}]
{
You are an expert analyst for a memory-augmented QA system. Analyze the failure cases below to identify why the system failed and how the memory management skills should change. \\

\#\# How This System Works \\
1. **Memory Storage**: The system applies memory management skills to decide what information to store from the text chunk. \\
2. **Memory Retrieval**: At question time, it retrieves the most relevant memories by semantic similarity. \\
3. **Answer Generation**: An LLM answers using the retrieved memories. \\

Failures can occur at any stage: \\
- **Storage failure**: Important information was never stored (skill missing or misapplied) \\
- **Retrieval failure**: Relevant memory exists but was not retrieved (embedding mismatch) \\
- **Memory quality failure**: Memory exists but is too vague or incomplete to answer \\

\#\# Current Memory Management Skills \\
\{operation\_bank\_description\} \\

\#\# Operation Evolution Feedback \\
\{evolution\_feedback\} \\

\#\# Failure Cases (\{num\_failure\_cases\} cases) \\
\{failure\_cases\_details\} \\

\#\# Analysis Instructions \\
This is round 1 of a reflection loop. Produce a strong initial analysis that can be critiqued and improved. \\
1. For each case, check whether the retrieved memories contain the answer or the needed evidence. \\
2. If missing, decide whether it was never stored (storage failure) or stored but too weak (memory quality failure). \\
3. If the answer is present but not retrieved, label it retrieval failure and avoid changing skills unless the pattern repeats. \\
4. Group cases into patterns tied to information types, entities, temporal details, or constraints. \\
5. For each pattern, propose a concrete skill change: add a new skill or refine an existing one to capture missing details. \\
6. Provide up to \{max\_changes\} recommendations total (use fewer if only one change is justified). \\

\#\# Output Format \\
Provide your analysis as JSON: \\
\{
    ``failure\_patterns": [
        \{
            ``pattern\_name": ``[descriptive name for this failure pattern]",
            ``affected\_cases": [list of case numbers, e.g., 1, 3, 5],
            ``root\_cause": ``[storage\_failure|retrieval\_failure|memory\_quality\_failure]",
            ``explanation": ``[why this pattern of failures is occurring]",
            ``potential\_fix": ``[what kind of operation change could address this]"
        \}
    ],
    ``recommendations": [
        \{
            ``action": ``[add\_new\_operation|refine\_existing\_operation|no\_change]",
            ``target\_operation": ``[operation name to refine, or null if adding new]",
            ``rationale": ``[clear explanation of why this is the best improvement]",
            ``priority": ``[high|medium|low]"
        \}
    ],
    ``summary": ``[1-2 sentence summary of main findings]"
\} \\

Focus on actionable insights. What specific change to the skill bank would prevent these failures?

Output ONLY the JSON, no other text.
}
\label{appendix:case_designer_1}
\end{tcolorbox}
\end{center}

\begin{center}
\begin{tcolorbox}[title={Designer Refinement Prompt (1/2)}]
{
Based on the failure analysis, propose a specific improvement to the memory operation bank. \\

\#\# Failure Analysis (from Stage 1) \\
\{analysis\_feedback\} \\

\#\# Current Operation Bank \\
\{operation\_bank\_full\} \\

\{evolution\_feedback\} \\

\#\# Your Task \\
Propose up to \{max\_changes\} improvements based on the analysis: \\

**Option A - Add New Operation**: Create a new operation if the analysis shows a capability gap (e.g., certain information types are not being captured). \\
**Option B - Refine Existing Operation**: Improve an existing operation's instruction template if the analysis shows it's not working well (e.g., memories are too vague, missing key details). \\
**Option C - No Change**: If the failures are due to retrieval issues (not operation issues), or if the current operations are already well-designed. \\

\#\# CRITICAL Requirements \\
1. instruction\_template MUST be a skill-style guide and MUST NOT include context placeholders (the executor injects the text chunk and retrieved memories) \\
2. instruction\_template MUST clearly state purpose, when to use, and constraints \\
3. instruction\_template MUST specify the allowed action type (INSERT or UPDATE only) \\
4. For new operations, `update\_type` must be either ``insert" or ``update" (delete and noop operations are not evolved at this time) \\
5. Only propose operations with update\_type ``insert" or ``update" \\
6. Avoid labels like ``ENHANCED", ``ADVANCED", or other marketing adjectives in descriptions or templates; keep phrasing neutral and task-specific \\
7. Do NOT embed output blocks; the executor handles output formatting and can apply the skill multiple times \\
8. The number of changes in the list MUST be less than {max\_changes} \\
9. Do NOT modify the same operation more than once in a single response, and do NOT refine an operation you add in the same response \\

\#\# Example of a Well-Designed Insert Operation \\
\{
    ``name": ``extract\_personal\_preferences",
    ``description": ``Memory management skill for capturing personal preferences and habits mentioned in the text chunk.",
    ``update\_type": ``insert",
    ``instruction\_template": ``Skill: Insert Preferences. Purpose: Capture personal preferences, habits, or opinions stated in the text chunk. When to use: The text chunk mentions likes, dislikes, routines, or goals tied to a person. How to apply: Attribute the preference to the correct person. Keep the preference specific and actionable. Constraints: Avoid one-off or ambiguous statements. Action type: INSERT only."
\} \\

\#\# Output Format \\
Respond with ONE of these JSON structures:

\#\#\# One or more changes (up to {max\_changes}):
\{
    ``action": ``apply\_changes",
    ``summary": ``[overall rationale for the set of changes]",
    ``changes": [
        \{
            ``action": ``add\_new",
            ``new\_operation": \{
                ``name": ``[snake\_case\_name]",
                ``description": ``[what it does and when to use it]",
                ``instruction\_template": ``[skill-style instruction template]",
                ``update\_type": ``[insert|update]",
                ``reasoning": ``[how this addresses the identified failures]"
            \}
        \},
        \{
            ``action": ``refine\_existing",
            ``refined\_operation": \{
                ``name": ``[existing\_operation\_name]",
                ``changes": \{
                    ``description": ``[improved description]",
                    ``instruction\_template": ``[improved template]"
                \},
                ``reasoning": ``[how these changes address the identified failures]"
            \}
        \}
    ]
\}

}
\label{appendix:case_designer_2_1}
\end{tcolorbox}
\end{center}

\begin{center}
\begin{tcolorbox}[title={Designer Refinement Prompt (2/2)}]
{
\#\#\# No changes needed: \\
\{
    ``action": ``no\_change",
    ``reasoning": ``[why the current operations are sufficient]"
\} \\

\#\# Instruction Template Structure \\
When writing instruction templates, follow this structure: \\

Skill: [Short skill name] \\
Purpose: [What this skill does] \\
When to use: \\
- [Trigger 1] \\
- [Trigger 2] \\
How to apply: \\
- [Step 1] \\
- [Step 2] \\
Constraints: \\
- [What to avoid] \\
Action type: [INSERT only | UPDATE only] \\

Output ONLY the JSON, no other text.
}
\label{appendix:case_designer_2_2}
\end{tcolorbox}
\end{center}

\begin{center}
\begin{tcolorbox}[title={LLM Judge Prompt}]
{
You are an expert judge evaluating the quality of an answer for a QA task.
Your goal is to determine whether the model's answer correctly and sufficiently
answers the given question. \\

Read the following information carefully: \\

[Question] \\
\{question\} \\

[Ground Truth Answers] \\
\{ground\_truth\} \\

[Model Answer] \\
\{model\_answer\} \\

Your evaluation criteria: \\
1. Correctness: \\
   - Is the model answer factually consistent with ANY of the correct answers? \\
   - Does it avoid contradictions or introducing false information? \\

2. Relevance: \\
   - Does the answer address the question directly without unnecessary content? \\

3. Completeness: \\
   - Does the answer include all essential information needed to fully answer the question? \\
   - Partial answers are allowed but should receive lower scores. \\

Scoring Rules: \\
- Score = 1.0 if the answer is fully correct. \\
- Score = 0.5 if the answer is partially correct but incomplete or slightly inaccurate. \\
- Score = 0.0 if the answer is incorrect, irrelevant, or contradicts the ground truth. \\

Output Format (STRICT): \\
Return your output as a JSON dictionary with two fields: \\
\{
    ``explanation": ``\textless brief explanation of your reasoning\textgreater",
    "score": \textless0.0 | 0.5 | 1.0\textgreater
\} \\

Be concise and objective. Do not include anything outside the JSON.
}
\label{appendix:case_judge}
\end{tcolorbox}
\end{center}

\section{Use of Large Language Models}
\label{app:llm_usage}

LLMs are used as part of our method and evaluation pipeline. Specifically, MemSkill uses an LLM executor for skill-conditioned memory generation, an LLM designer for skill evolution, and an LLM judge for evaluation on tasks where automatic metrics are insufficient. The corresponding prompts are provided in Appendix~\ref{appendix:prompts}.

LLM-based tools may also be used for grammar polishing and code assistance. The authors take full responsibility for all scientific claims, experimental results, figures, and references in this paper.

\section{Limitations and Societal Impact}
\label{app:limitations_impact}

\paragraph{Limitations.}
This work focuses on benchmarked research settings for skill-conditioned memory construction. Practical deployments of agent memory systems may require additional mechanisms for privacy protection, user consent, access control, and data retention. These considerations are orthogonal to the core algorithmic contribution of MemSkill and are left for future deployment-oriented studies.

\paragraph{Societal impact.}
MemSkill contributes toward more adaptive and reusable memory mechanisms for long-horizon LLM agents. Instead of repeatedly designing task-specific memory pipelines, the proposed skill-based formulation allows memory behaviors to be learned, reused, and evolved from interaction traces, which may lower the engineering barrier for building memory-augmented agents. This can benefit research and practical applications that require long-term context management, such as personalized assistants, educational agents, research support tools, and embodied task-solving systems. Since agent memory may involve user-provided or task-specific information, practical deployments should follow standard data governance practices, including user consent, privacy protection, data retention control, and mechanisms for inspecting or deleting stored memories. Our work focuses on benchmarked research settings and does not target surveillance, deception, or high-stakes decision making.

%%%%%%%%%%%%%%%%%%%%%%%%%%%%%%%%%%%%%%%%%%%%%%%%%%%%%%%%%%%%

\newpage
\section*{NeurIPS Paper Checklist}

\begin{enumerate}

\item {\bf Claims}
    \item[] Question: Do the main claims made in the abstract and introduction accurately reflect the paper's contributions and scope?
    \item[] Answer: \answerYes{} % Replace by \answerYes{}, \answerNo{}, or \answerNA{}.
    \item[] Justification: The abstract and introduction summarize the proposed skill-conditioned memory construction formulation, the main components of MemSkill, and the empirical settings used to support the claims.

\item {\bf Limitations}
    \item[] Question: Does the paper discuss the limitations of the work performed by the authors?
    \item[] Answer: \answerYes{} % Replace by \answerYes{}, \answerNo{}, or \answerNA{}.
    \item[] Justification: We discuss the limitations of MemSkill in Appendix~\ref{app:limitations_impact}, including benchmarked research settings and practical considerations such as privacy, user consent, access control, and data retention.

    \item {\bf Theory assumptions and proofs}
    \item[] Question: For each theoretical result, does the paper provide the full set of assumptions and a complete (and correct) proof?
    \item[] Answer: \answerNA{} % Replace by \answerYes{}, \answerNo{}, or \answerNA{}.
    \item[] Justification: The paper does not include formal theoretical results. We provide the RL training objective and optimization formulation in Appendix~\ref{appendix:rl_objective}.

    \item {\bf Experimental result reproducibility}
    \item[] Question: Does the paper fully disclose all the information needed to reproduce the main experimental results of the paper to the extent that it affects the main claims and/or conclusions of the paper (regardless of whether the code and data are provided or not)?
    \item[] Answer: \answerYes{} % Replace by \answerYes{}, \answerNo{}, or \answerNA{}.
    \item[] Justification: We provide the implementation details, model configurations, prompts, training objective, dataset settings, evaluation protocol, and hyperparameters needed to reproduce the main experiments in Section~\ref{sec:exp} and~\ref{appendix:imple_detail}.

\item {\bf Open access to data and code}
    \item[] Question: Does the paper provide open access to the data and code, with sufficient instructions to faithfully reproduce the main experimental results, as described in supplemental material?
    \item[] Answer: \answerNo{} % Replace by \answerYes{}, \answerNo{}, or \answerNA{}.
    \item[] Justification: We do not provide codes at submission time to preserve anonymity, but the paper and appendix include detailed implementation details, prompts, dataset settings, hyperparameters, and evaluation protocols needed to reproduce the main results (Section~\ref{sec:exp} and~\ref{appendix:imple_detail}). We will fully release the code upon publication.

\item {\bf Experimental setting/details}
    \item[] Question: Does the paper specify all the training and test details (e.g., data splits, hyperparameters, how they were chosen, type of optimizer) necessary to understand the results?
    \item[] Answer: \answerYes{} % Replace by \answerYes{}, \answerNo{}, or \answerNA{}.
    \item[] Justification: The paper specifies the datasets and splits, model backbones, training setup, optimizer and hyperparameters, retrieval settings, prompts, and evaluation protocols in Section~\ref{sec:exp} and~\ref{appendix:imple_detail}.

\item {\bf Experiment statistical significance}
    \item[] Question: Does the paper report error bars suitably and correctly defined or other appropriate information about the statistical significance of the experiments?
    \item[] Answer: \answerNo{} % Replace by \answerYes{}, \answerNo{}, or \answerNA{}.
    \item[] Justification: We fix the seed for all our experiments, which ensures the reliability of our experiments.

\item {\bf Experiments compute resources}
    \item[] Question: For each experiment, does the paper provide sufficient information on the computer resources (type of compute workers, memory, time of execution) needed to reproduce the experiments?
    \item[] Answer: \answerYes{} % Replace by \answerYes{}, \answerNo{}, or \answerNA{}.
    \item[] Justification: We describe the compute resources used for the experiments in Section~\ref{sec:exp} and Appendix~\ref{appendix:imple_detail}.
    
\item {\bf Code of ethics}
    \item[] Question: Does the research conducted in the paper conform, in every respect, with the NeurIPS Code of Ethics \url{https://neurips.cc/public/EthicsGuidelines}?
    \item[] Answer: \answerYes{} % Replace by \answerYes{}, \answerNo{}, or \answerNA{}.
    \item[] Justification: The research follows the NeurIPS Code of Ethics, uses standard public benchmarks and evaluation protocols, preserves submission anonymity, and does not involve human-subject experiments, sensitive personal data collection, or high-stakes decision making.

\item {\bf Broader impacts}
    \item[] Question: Does the paper discuss both potential positive societal impacts and negative societal impacts of the work performed?
    \item[] Answer: \answerYes{} % Replace by \answerYes{}, \answerNo{}, or \answerNA{}.
    \item[] Justification: We discuss broader impacts in Appendix~\ref{app:limitations_impact}.
    
\item {\bf Safeguards}
    \item[] Question: Does the paper describe safeguards that have been put in place for responsible release of data or models that have a high risk for misuse (e.g., pre-trained language models, image generators, or scraped datasets)?
    \item[] Answer: \answerNA{} % Replace by \answerYes{}, \answerNo{}, or \answerNA{}.
    \item[] Justification: The paper poses no such risks.

\item {\bf Licenses for existing assets}
    \item[] Question: Are the creators or original owners of assets (e.g., code, data, models), used in the paper, properly credited and are the license and terms of use explicitly mentioned and properly respected?
    \item[] Answer: \answerYes{} % Replace by \answerYes{}, \answerNo{}, or \answerNA{}.
    \item[] Justification: We use several existing datasets in our work and have ensured that all are properly credited. 

\item {\bf New assets}
    \item[] Question: Are new assets introduced in the paper well documented and is the documentation provided alongside the assets?
    \item[] Answer: \answerNA{} % Replace by \answerYes{}, \answerNo{}, or \answerNA{}.
    \item[] Justification: The paper does not release new assets.

\item {\bf Crowdsourcing and research with human subjects}
    \item[] Question: For crowdsourcing experiments and research with human subjects, does the paper include the full text of instructions given to participants and screenshots, if applicable, as well as details about compensation (if any)? 
    \item[] Answer: \answerNA{} % Replace by \answerYes{}, \answerNo{}, or \answerNA{}.
    \item[] Justification: The paper does not involve crowdsourcing nor research with human subjects.

\item {\bf Institutional review board (IRB) approvals or equivalent for research with human subjects}
    \item[] Question: Does the paper describe potential risks incurred by study participants, whether such risks were disclosed to the subjects, and whether Institutional Review Board (IRB) approvals (or an equivalent approval/review based on the requirements of your country or institution) were obtained?
    \item[] Answer: \answerNA{} % Replace by \answerYes{}, \answerNo{}, or \answerNA{}.
    \item[] Justification: The paper does not involve crowdsourcing nor research with human subjects.

\item {\bf Declaration of LLM usage}
    \item[] Question: Does the paper describe the usage of LLMs if it is an important, original, or non-standard component of the core methods in this research? Note that if the LLM is used only for writing, editing, or formatting purposes and does \emph{not} impact the core methodology, scientific rigor, or originality of the research, declaration is not required.
    %this research? 
    \item[] Answer: \answerYes{} % Replace by \answerYes{}, \answerNo{}, or \answerNA{}.
    \item[] Justification: We describe the use of LLMs in Appendix~\ref{app:llm_usage}, including their roles as the executor for skill-conditioned memory generation, the designer for skill evolution, and the judge for evaluation.

\end{enumerate}

\end{document}